\documentclass[letterpaper, 10 pt, journal, twoside]{IEEEtran}  

\IEEEoverridecommandlockouts                              

\usepackage{amsmath} 
\usepackage{amssymb}  
\usepackage[utf8]{inputenc}
\usepackage[keeplastbox]{flushend}
\usepackage{graphicx}
\usepackage{cite}
\usepackage{siunitx}
\usepackage[caption=false]{subfig}
\usepackage{afterpage}
\usepackage{booktabs}
\usepackage{siunitx}
\usepackage{multicol}
\usepackage{mathdots} 

\title{PHASER: a Robust and Correspondence-free Global Pointcloud Registration
\author{Lukas Bernreiter, Lionel Ott, Juan Nieto, Roland Siegwart and Cesar Cadena} 
\thanks{Manuscript received: September 27, 2020; Revised December 3, 2020;
Accepted January 2, 2021.}
\thanks{This paper was recommended for publication by Editor Sven Behnke upon evaluation of the Associate Editor and Reviewers' comments.}
\thanks{This work was supported by the National Center of Competence in Research (NCCR) Robotics through the Swiss National Science Foundation and has received funding from the European Union’s Horizon 2020 research and innovation programme under grant agreement No 871542.}
\thanks{All authors are with the Autonomous Systems Lab, ETH Zurich, Zurich 8092, Switzerland, {\tt \small \{berlukas, lioott, nietoj, rsiegwart, cesarc\}@ethz.ch.}}%
\thanks{Digital Object Identifier (DOI): see top of this page.}
}

\usepackage[utf8]{inputenc}
\usepackage[OT1]{fontenc}

\usepackage{fancyhdr}

\usepackage{textcomp}\usepackage{gensymb}

\usepackage{multicol}

\usepackage{subfig}
\usepackage{graphicx}

\usepackage{booktabs}
\usepackage{array}
\usepackage{multirow}

\usepackage{color}
\usepackage{colortbl}
\definecolor{black}{rgb}{0,0,0}
\definecolor{white}{rgb}{1,1,1}
\definecolor{darkred}{rgb}{0.5,0,0}
\definecolor{darkgreen}{rgb}{0,0.5,0}
\definecolor{darkblue}{rgb}{0,0,0.5}

\usepackage{amsmath}
\usepackage{amssymb}

\usepackage{nicefrac}

\usepackage{upgreek}

\usepackage{isomath}
\renewcommand{\vec}{\vectorsym}
\newcommand{\mat}{\matrixsym}

\usepackage{units}

\usepackage{rotating}

\usepackage{pdfpages}
\includepdfset{pages={-}, frame=true, pagecommand={\thispagestyle{fancy}}}

\usepackage{hyperref}

\usepackage{cleveref}

\usepackage[nolist,nohyperlinks]{acronym}

\newcommand{\norm}[1]{\left\lVert#1\right\rVert}

\renewcommand{\vec}[1]{\ensuremath{{\boldsymbol{#1}}}}

\begin{acronym}
\acro{SLAM}{Simultaneous Localization And Mapping}
\acro{TSDF}{Truncated Signed Distance Field}
\acro{ICP}{Iterative Closest Point}
\acro{RFS}{Random Finite Sets}
\acro{CNN}{Convolutional Neural Network}
\end{acronym}

\begin{document}
\maketitle
%
\begin{abstract}
We propose PHASER, a correspondence-free global registration of sensor-centric pointclouds that is robust to noise, sparsity, and partial overlaps. 
Our method can seamlessly handle multimodal information, and does not rely on keypoint nor descriptor preprocessing modules.
By exploiting properties of Fourier analysis, PHASER operates directly on the sensor's signal, fusing the spectra of multiple channels and computing the 6-DoF  transformation based on correlation. 
Our registration pipeline starts by finding the most likely rotation $\vec{r}\in{}SO(3)$ followed by computing the most likely translation $\vec{t}\in{}\mathbb{R}^3$.
Both estimates, $\vec{r}$, and $\vec{t}$ are distributed according to a probability distribution that takes the underlying manifold into account, i.e., a Bingham and a Gaussian distribution, respectively. 
This further allows our approach to consider the periodic-nature of $\vec{r}$ and naturally represents its uncertainty.
We extensively compare PHASER against several well-known registration algorithms on both simulated datasets, and real-world data acquired using different sensor configurations.
Our results show that PHASER can globally align pointclouds in less than $100\,\mathrm{ms}$ with an average accuracy of $2\,\mathrm{cm}$ and $0.5^\circ$, is resilient against noise, and can handle partial overlap.
\end{abstract}
\begin{IEEEkeywords}
 Mapping, Sensor Fusion, Probability and Statistical Methods
\end{IEEEkeywords}
\IEEEpeerreviewmaketitle
\section{Introduction}
\IEEEPARstart{T}{he} registration of pointclouds from  3D scanners stems as a core competence for numerous applications in robotics such as state estimation, map building, and localization~\cite{Zhang, Pomerleau2015}.
Especially since recent advances of more affordable and effective RGB-D and LiDAR scanners leverage to infer an environment's geometry readily. 
It is further particularly crucial for ill-lighted and unstructured environments, where LiDAR systems constitute the preferred sensing option, as they are inherently invariant to illumination changes yielding a potentially more reliable robotics system.

Traditional local approaches such as \ac{ICP} are highly sensitive to local minima in the objective function, making a good prior initialization of the registration an indispensable requirement.
However, often in robotics, only a very poor initialization is available. Therefore, global registration systems have gained increased interest since they can provide an optimal solution without requiring an initial guess. 
Many of the existing global registration approaches rely on computing point correspondences by matching unique point descriptors between the scans~\cite{Rusu2009} stating that a subset of points from the source scan corresponds to another in the target scan.
However, relying on such point descriptors raises the concern that an algorithm's efficiency and success depend highly on the extracted descriptor's quality.
Especially in the context of localization, where the goal is to find a transformation between a local scan to a prior map, this is inherently flawed since the points are distorted and noisy.
Hence, they perturb the extraction of the descriptors and consequently the matching. 
Furthermore, the global alignment problem becomes significantly more challenging when the data contains different viewpoints yielding confined overlap between the pointclouds and many outlier correspondences.
Alternatively, approaches which do not rely on correspondences are less affected by such issues and can potentially increase the overall system's robustness~\cite{Le2019}.

Also, it is evident that in many of the aforementioned cases, an assessment of the alignment quality is another critical aspect for several robotic applications, for instance, its uncertainty for incorporating loop-closure factors in graph-based SLAM.
However, finding the proper balance between the computation time and the uncertainty's precision is considerably challenging and remains an active research field~\cite{Landry2019}.

Moreover, conventional global pointcloud registration systems often neglect the additional channels and information a LiDAR sensor provides.
Besides range information, LiDAR sensors typically also return the received energy level (intensity) and the ambient IR. 
Both modalities represent additional characteristics of the reflected surface and provide valuable information about the environment but need careful consideration when incorporated into a registration pipeline.
%
\begin{figure}[!t]
\centering
    \subfloat[Noisy pointclouds] {%
        \includegraphics[width=0.40\columnwidth, trim={4.5cm 7.5cm 4.5cm 7.5cm}, clip]{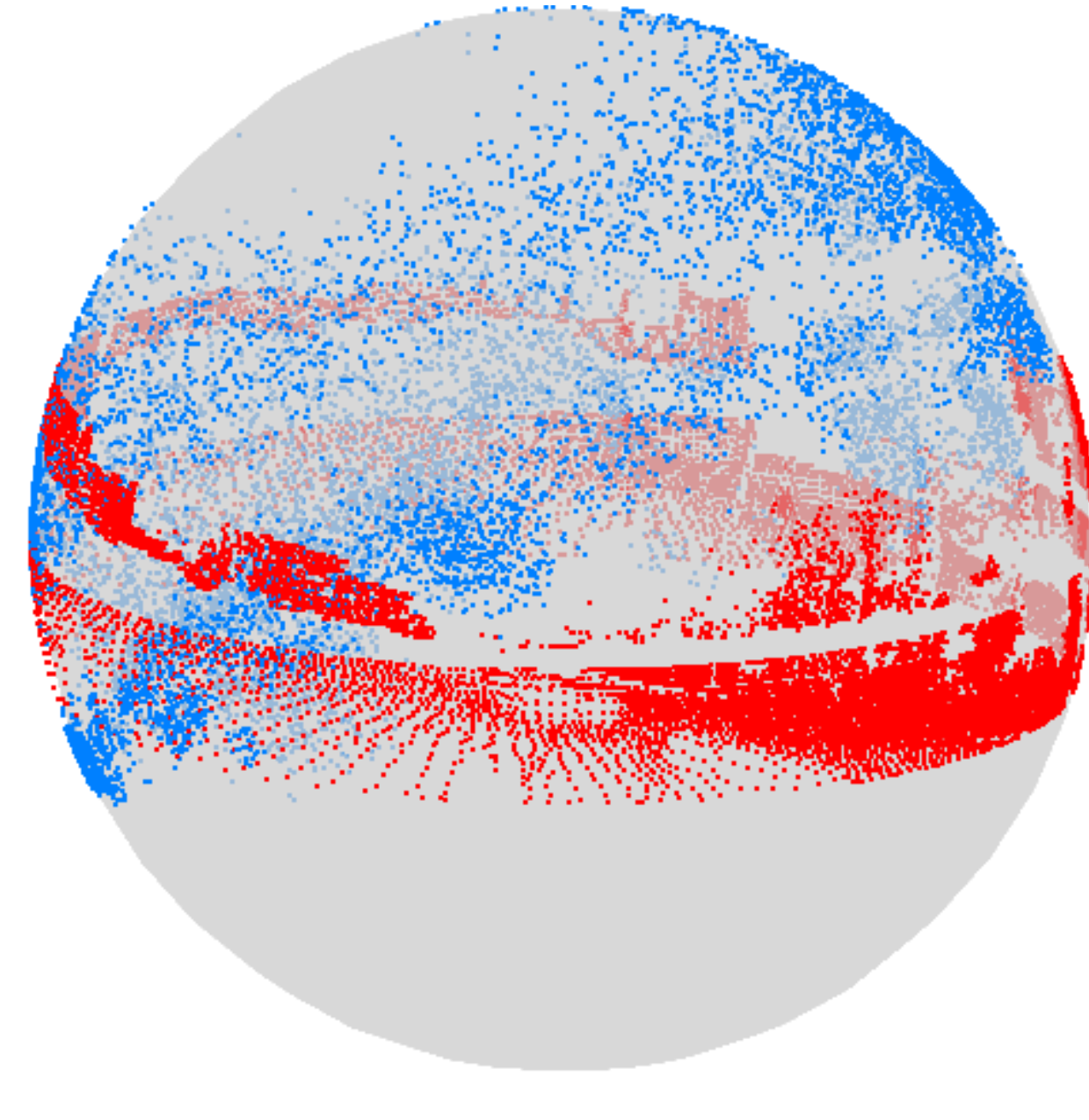}%
        \label{fig:teaser:input}
    } \hspace{1cm}
    \subfloat[Globally registered] {%
        \includegraphics[width=0.4\columnwidth, trim={4.5cm 7.5cm 4.2cm 7.5cm}, clip]{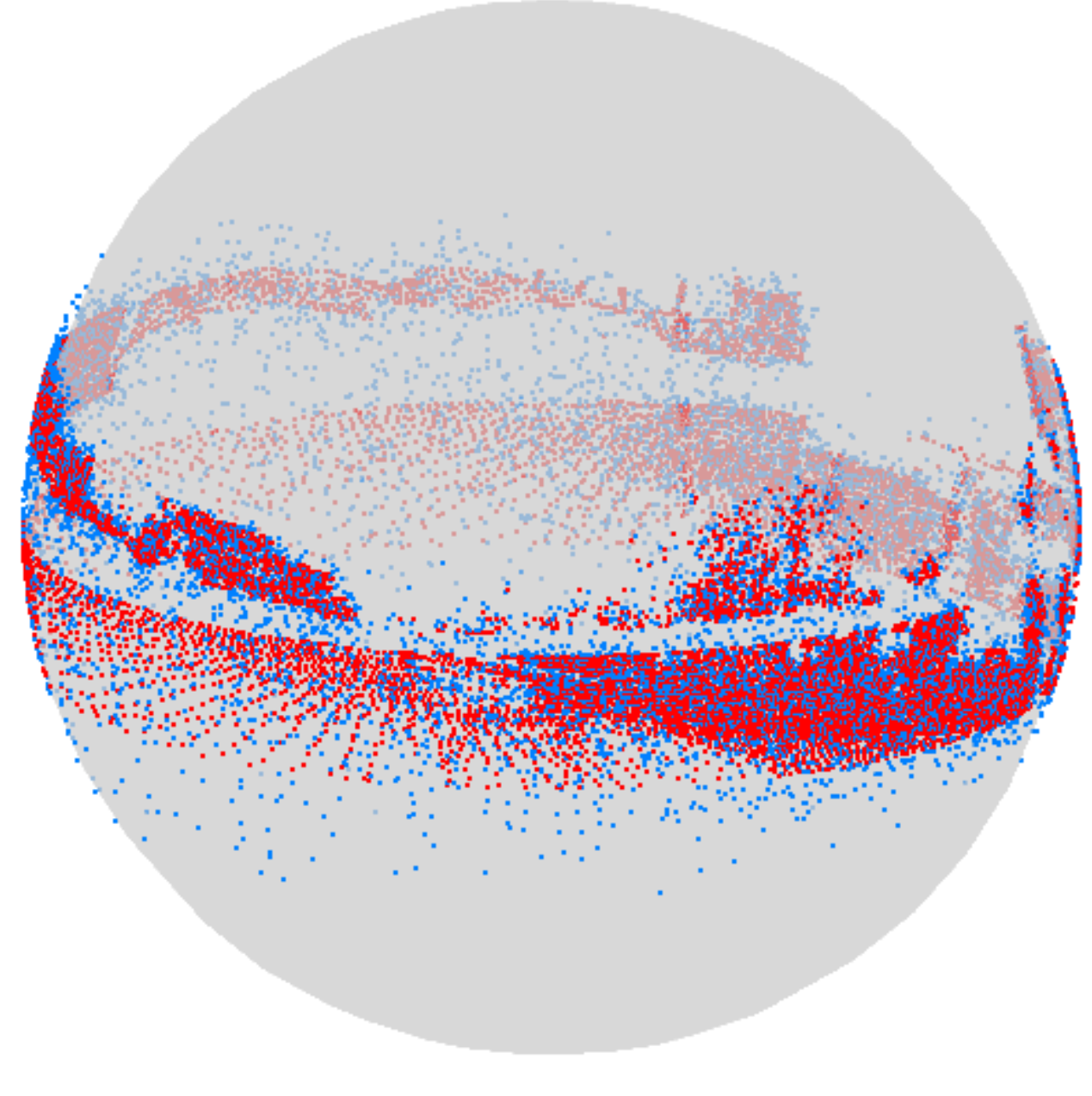}
        \label{fig:teaser:reg}
    }
    \caption{We propose a global registration pipeline by employing a Fourier analysis on the pointclouds. Given two pointclouds in (a), we find the correct global alignment for the rotation and translation, as in (b).}
    \label{pics:teaser}
    \vspace{-5mm}
\end{figure}
\begin{figure*}[!t]
  \vspace{0.2cm}
  \centering
   \includegraphics[width=0.95\textwidth, trim={0.0cm, 0cm, 0.0cm, 0cm}, clip]{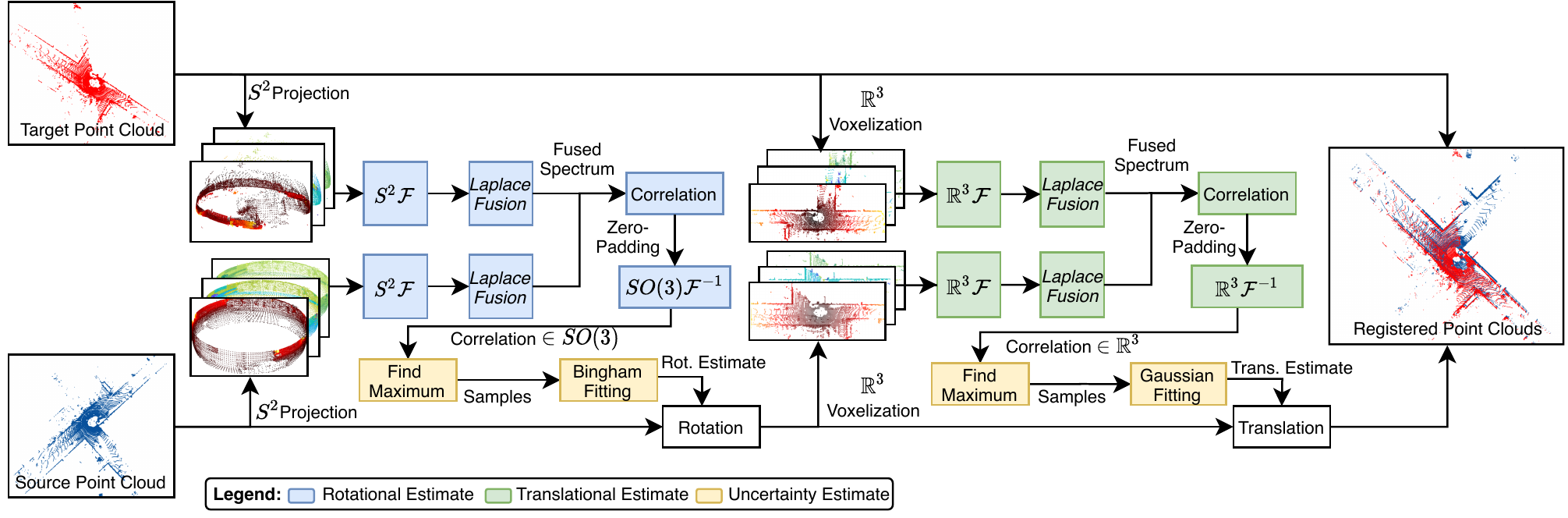}
   \vspace{-3mm}
   \caption{Overview of the proposed system. For the registration core, we project both, source and target pointcloud onto the sphere and perform a spherical Fourier transform ($S^2\mathcal{F}$) to find the rotation that maximizes the spherical correlation (blue). Using the oriented scans, we estimate the translation as a next step using a spatial Fourier transform ($\mathbb{R}^3\mathcal{F}$) and correlation (green). Our framework additionally examines each correlation's result to infer a measure of the registration uncertainty (yellow). If additional modalities are used, our pipeline will automatically fuse their spectra using a Laplacian pyramid before the correlation step is performed.}
   \label{pics:method:overview}
   \vspace{-7mm}
\end{figure*}

In this paper we propose a solution based on Fourier analysis. The approach transforms the pointclouds to the frequency domain, where the frequency components of multiple modalities are fused and then aligned through correlation. 
As a consequence, our approach does not require any point correspondences, initial guess, nor additional parameters for the fusion.
We make use of a spherical projection model (cf.  Figure~\ref{pics:teaser}) to enable a decomposition into spherical frequencies of the pointclouds.
Furthermore, we derive a measure for the registration uncertainty using the outcome of the correlation.
In more detail, for the translation we fit a multivariate Gaussian distribution to the resulting correlation to estimate the spatial uncertainty. 
For the rotation, we make use of directional statistics to account for the fact that the estimate does not lie in the Euclidean space but rather on a curved manifold.
Specifically, we use the Bingham distribution which inherently takes the periodic nature of the rotation into account. 
A general overview of our registration pipeline is given in Figure~\ref{pics:method:overview}. 
We denote our proposed registration pipeline as {\em{PHASER}}, short for phase registration.

The resulting global registration pipeline is evaluated in simulations where we can verify the correctness of the algorithm, and in real-world experiments to show its pragmatic applicability. 
The main contributions of this work are:
\begin{itemize}
    \item A spherical and spatial Fourier decomposition of band-limited signals to individually estimate the rotation and translation of pointclouds. 
    \item A fusion methodology for arbitrary input modalities. 
    \item A method for estimation of the spherical and spatial correlation uncertainty using a Bingham and a Gaussian distribution, respectively.
\end{itemize}
%
\section{Related Work}
\label{sec:related_work}
Pointcloud registration is a well studied field with a variety of different applications and there is extensive literature available, e.g. Pomerleau~\cite{Pomerleau2015} provides a good overview in the context of robotics. 
Typically, registration approaches can be divided into local and global approaches and further can rely on correspondences between the pointclouds or operate correspondence-free. 
\subsection{Local Pointcloud Registration}
If the correspondences are known it was shown in the seminal work of Horn~\cite{Horn1987} that registration can be solved in closed form.
For unknown correspondences, traditional approaches such as \ac{ICP} iteratively refine the correspondences and compute the optimal solution using Horn's method~\cite{Horn1987} at each iteration.
Consequently, the registration quality substantially decreases in the presence of noise and motion-distortions.
Therefore, local methods as \ac{ICP} strongly depend on a good initial prior. 
To overcome some robustness issues, probabilistic approaches such as NDT~\cite{Magnusson2007}, Generalized-ICP~\cite{Segal2009} and the GMM-based registration~\cite{Tabib2018} fit a Gaussian distribution to the points or voxels and optimize over a distribution-to-distribution distance.  

While the majority relies on geometric information, there has been some work which includes additional information such as color or intensity to e.g. reject outliers~\cite{Parkison2019} or partition distributions~\cite{Zaganidis2018}.
Compared to all previously mentioned methods, ours does not minimize a metrical error in vector space but rather measures the similarity between clouds and therefore, operates globally.
\subsection{Global Pointcloud Registration}
Global pointcloud registration approaches, such as the work of Rusu et al.~\cite{Rusu2009} and Lei et al.~\cite{Lei2017} seek to efficiently find a unique solution by extracting point descriptors to find correspondences for the registration without prior knowledge.
The work Le et al.~\cite{Le2019} proposes a randomized approach to overcome issues with computing descriptors and point correspondences.
Yang et al.~\cite{Jiaolong2015} proposed a global variant of the \ac{ICP} algorithm based on the branch-and-bound technique for global optimization.

Yang and Carlone~\cite{Yang2020} approach the motion estimation by decoupling the rotation from the translation estimate. 
Their method targets the case of extreme outlier correspondences and shows increased robustness against these.
Making use of the Fourier domain also enables decoupling rotation from translation which is based on the fact that phase shifts do not affect the magnitude of a transformed signal~\cite{Bulow2013,Bulow2018}. 
Correlation-based registration approaches in the Fourier domain have shown good robustness for finding translations in 2D~\cite{SrinivasaReddy1996} and 3D~\cite{Wang2012} and further, do not require computing correspondences. 
Bülow et al.~\cite{Bulow2013} propose a heuristic-based resampling of FFT-layers to compute the rotational difference while the translational offset is found by matching the phases.
Generalizing the Fourier transform to the spherical domain additionally enables finding the orientation between pointclouds~\cite{Bulow2018, Makadia2006}.
Bülow and Birk~\cite{Bulow2018} proposed an extension to the well known Fourier-Mellin transform to include a spherical correlation resulting in a pipeline for estimating rotation, translation, and scale.
Further, Makadia et al. proposed a full registration pipeline for pointclouds~\cite{Makadia2006} utilizing orientation histograms of normals and a convolution over the occupancy.
Likewise, our method's registration core also separates rotation from translation and transforms the input using a spherical and spatial Fourier transform, respectively.
In contrast to~\cite{Bulow2018, Makadia2006}, our approach encodes more characteristics such as signal intensities instead of only occupancy information.

Many of the mentioned global pointcloud registration approaches are often tailored to 3D scanned objects with uniformly distributed points or to provide only a coarse alignment which is later fine-tuned by \ac{ICP}. 
Notably, LiDAR scans in robotics are often distorted, have a large, irregular spatial extent, and are affected by noise.
Thus, require a robust registration methodology and an evaluation of its quality.
Our method can deal with high noise levels by operating in the spherical and spatial Fourier domain where we fuse and correlate the spectra of multiple modalities using a Laplacian pyramid.
We can further configure our pipeline for a coarse or fine registration and, using the correlation, approximate the registration uncertainty.
\subsection{Registration Uncertainty}
Landry et al.~\cite{Landry2019}. present an overview of current approaches for uncertainty estimation for registration.
The solutions can be grouped into two main groups: sampling-based approaches~\cite{Nieto2006} and closed-form solutions~\cite{Prakhya2015}. 
Typically, sampling-based approaches have high computational demand and often do not scale well with large search spaces. 
In contrast, closed-form solutions are often less computational-intensive but should be used cautiously when noise is present~\cite{Landry2019}. 
The solution presented in this paper differs by requiring only a small amount of weighted correlation samples, which are then fitted to a probability distribution.
%
\section{Preliminaries}
\label{sec:preliminaries}
We address the registration problem by performing a correlation between source and target pointcloud in the Fourier domain.
Thus, we first introduce the necessary concepts needed to perform a correlation in the spherical and spatial Fourier domain.
This essentially covers the theory of harmonic analysis on the rotation group as well as on the vector space. 
Further, we introduce the directional distribution used for the orientation uncertainty.
\subsection{Spatial and Spherical Fourier Analysis}\label{sec:pre:fourier}
In general, the spatial Fourier transform of a function $f\in\mathbb{R}^3$ is defined as the inner product $F(\vec{s}) = \langle{}f, \exp\left(j2\pi \vec{s}\right)\rangle$,
where $j$ is the imaginary unit and $\vec{s}=[u,v,w]^\top$ the spatial frequency coordinates.
Using $n^3$ equally spaced samples (i.e. voxels) over $\mathbb{R}^3$ leads to its well known discrete form, i.e.
\begin{align}\label{eq:pre:spatial_fourier}
    {F}(u,v,w) = \frac{1}{n^3}\sum_{x=0}^{n-1}&\sum_{y=0}^{n-1}\sum_{z=0}^{n-1} f(x,y,z)\\ \nonumber
    &\exp\left(-j\frac{2\pi}{n} [x,y,z]^\top[u,v,w]\right).
\end{align}
Generally, given the sensors' field of view $FoV$, the spatial resolution $\Delta=\frac{FoV}{n^3}$ (i.e. voxel width) directly controls the used bandwidth $B = \frac{2\pi}{\Delta}$ in the frequency domain and consequently the discretization of the spectrum.

Moreover, it is well known that the Fourier coefficients retain their phase information under translation of the original signal.
Thus, a translational shift in spatial coordinates results in a phase difference in the frequency domain which can be readily found using correlation techniques~\cite{SrinivasaReddy1996, Wang2012}.

Generally, the theory behind the Fourier analysis of signals can be extended to the hypersphere $S^2=\left\{x \in \mathbb{R}^3 \, | \, \norm{x}_2=1\right\}$, where we use the same parametrization as in Healy et al.~\cite{Healy2003}, i.e. $\vec{\omega}(\phi,\theta)=[\cos\phi\sin\theta, \sin\phi\sin\theta,\cos\theta]^\top$ for $\vec{\omega}\in S^2$, $\phi\in[0,2\pi]$ and $\theta\in[0,\pi]$.
The Hilbert space of $L^2(S^2)$ is defined over the inner product as
\begin{equation}\label{eq:pre:s2_integral}
    \langle{}\tilde{f},\tilde{h}\rangle=\int_{\vec{\omega}\in{}S^2} \tilde{f}(\vec{\omega})\overline{\tilde{h}(\vec{\omega})}d\vec{\omega}, 
\end{equation}
where $d\vec{\omega} = \sin(\theta)d\theta d\phi$, $\tilde{f},\tilde{h}$ are arbitrary square-integrable functions on $S^2$ and $\overline{\tilde{h}}$ its complex conjugate. 

The spherical Fourier transform of any function $\tilde{f}\in{}S^2$ is defined as the inner product $\tilde{F}_{m}^l = \langle{}\tilde{f}, Y_m^l\rangle$, where $Y_m^l$ are the so called spherical harmonics of degree $l\in\mathbb{N}_0$, order $m\in[0,l]\in\mathbb{N}_0$ and form an orthonormal basis over $L^2(S^2)$, i.e.
\begin{align}
    Y_m^l(\vec{\omega}) = (-1)^m\sqrt{\frac{(2l+1)(l-m)!}{4\pi (l+m)!}} P_m^l(\cos\theta)\exp(-jm\phi),
\end{align}
where $P_m^l$ are the associated Legendre polynomials which can be calculated numerically stable as proposed in the work by Healy et al.~\cite{Healy1996}.

We evaluate the integral over the sphere in eq.~\ref{eq:pre:s2_integral} using equiangular samples according to the sampling theorem by Driscoll and Healy~\cite{Driscoll1994} which results in the following Fourier coefficients:
\begin{equation}\label{eq:pre:spherical_fourier}
    \tilde{F}_{m}^l = \frac{\sqrt{2\pi}}{2\tilde{B}}\sum_{j=0}^{2\tilde{B}-1}\sum_{k=0}^{2\tilde{B}-1} w_j \tilde{f}(\theta_j,\phi_k)\overline{Y}_m^l(\theta_j,\phi_k),
\end{equation}
where $\tilde{B}$ is the spherical bandwidth, $\vec{\theta}$ and $\vec{\phi}$ are the samples with the corresponding weights $\vec{w}$.
Combining eq.~\eqref{eq:pre:spherical_fourier} to a set of coefficients results in the spherical Fourier transform for a given $\tilde{B}$. 

Furthermore, this enables the expansion of arbitrary functions $\tilde{f}\in L^2(S^2)$ in the base of the spherical harmonics, i.e.
\begin{equation}\label{eq:pre:fourier_expansion}
    \tilde{f}(\vec{\omega}) = \sum_{l\ge0}\sum_{m\le{}l} \tilde{F}_m^l Y_m^l(\vec{\omega}).
\end{equation}
Our orientation estimation for pointclouds is based on the fact that the magnitude $|\tilde{F}|$ is invariant under rotation~\cite{Kostelec2008}.
Thus, similar to the case of a spatial displacement, the rotational difference between two signals can be found using correlation techniques~\cite{Makadia2006, Bulow2018}. 
\subsection{Orientation Uncertainty Representation}
In this work we make use of the Bingham ditribution, a directional distribution, which naturally arises when conditioning a zero-mean Gaussian distribution on the hypersphere~\cite{Bingham1974}.
Its probability density function is defined as
\begin{equation}
    \mathcal{B}(\vec{x};\mat{M},\mat{Z})=\frac{1}{N(\mat{Z})}\exp{\left(\vec{x}^\top\mat{M}\mat{Z}\mat{M}^\top\vec{x}\right)}, 
\end{equation}
where $\mat{M}$ is the orientation, $\mat{Z}$ are the concentration parameters and $N(\mat{Z})$ a normalization constant.

The Bingham distribution is antipodal symmetric, takes the periodicity into account, and also naturally models the theory of unit-quaternions, i.e. the fact that $\vec{q}$ and $-\vec{q}$ represent the same rotation.
Consequently, an uncertain orientation can be modeled as a unit-quaternion that is distributed according to a 4D Bingham distribution.
With this formulation we can now utilize the second-order moment of a Bingham distribution as an uncertainty metric. It can be derived as follows
\begin{equation}\label{eq:simga_b}
    \Sigma_\mathcal{B} = \mat{M}\cdot \operatorname{diag}\left(\frac{\frac{\partial}{\partial z_1}N(\mat{Z})}{N(\mat{Z})}, ..., \frac{\frac{\partial}{\partial z_d}N(\mat{Z})}{N(\mat{Z})}\right) \mat{M}^\top,
\end{equation}
where $d$ is the dimension of the distribution.
\section{Method}
\label{sec:method}
We seek to find a rigid-body motion in $SE(3)$, with its uncertainty, which transforms a source to a target pointcloud.
In this section, we outline the core components of our pipeline shown in Figure~\ref{pics:method:overview}. Starting with the source and target clouds, this comprises: (i) input representation, (ii) orientation estimation, (iii) translation estimation and (iv) fusion of multiple modalities.
%
%
\subsection{Representative Functions}\label{sec:method:representation}
The input pointclouds are represented by two functions, i.e., $f,h\in\mathbb{R}^3$ and $\tilde{f},\tilde{h}\in{}S^2$.
Throughout this paper, we use $f\text{ and }\tilde{f}$ for the target, $h\text{ and }\tilde{h}$ for the source pointcloud.

Furthermore, we define a set of $K$ modalities suitable for correlation analysis, i.e. $f_i,h_i$, and $\tilde{f}_i,\tilde{h}_i$ for $i\in{}\{1,\dots,K\}$.
Generally, these functions can be chosen arbitrarily, although they ought to represent the environment somehow. Simple mappings of the individual channels (e.g., intensity, reflectivity), or handcrafted features (e.g., corner and surface features), are suitable choices.
However, we will first cover the registration core with a single modality ($K=1$) for simplicity and denote $f=f_i, h=h_i$ and $\tilde{f}=\tilde{f}_i,\tilde{h}=\tilde{h}_i$ before we discuss the fusion of different modalities.
The functions $f,h$ and $\tilde{f},\tilde{h}$ are transformed, fused, and correlated in the spatial and spherical frequency domain, respectively, to infer the relative transformation between them.
Since the translational part requires oriented pointclouds, we first perform the spherical correlation. 
\subsection{Rotation Estimation}\label{sec:method:rot}
\begin{figure}[!t]
\vspace{-2mm}
\centering
    \subfloat[OS1-64] {%
        \includegraphics[width=0.27\columnwidth, trim={7.05cm 11.5cm 6.85cm 11.2cm}, clip]{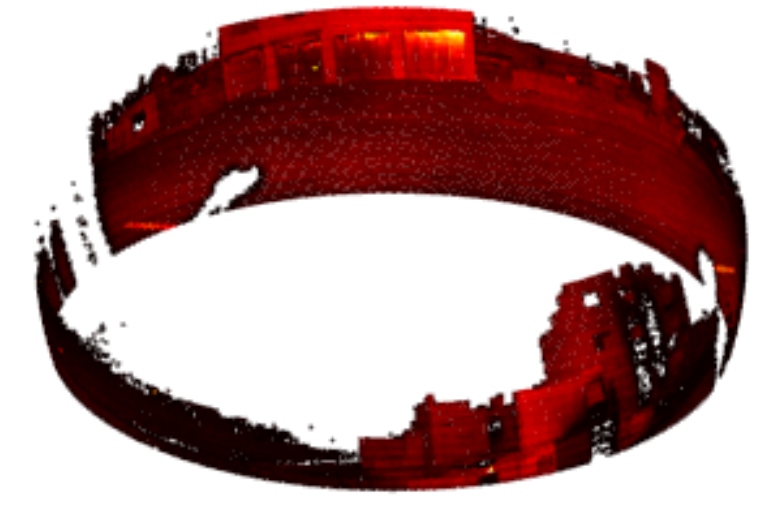}%
        \label{fig:os1}
    } \hfill
    \subfloat[RS-BPearl] {%
        \includegraphics[width=0.27\columnwidth, trim={7.05cm 11cm 6.85cm 11.5cm}, clip]{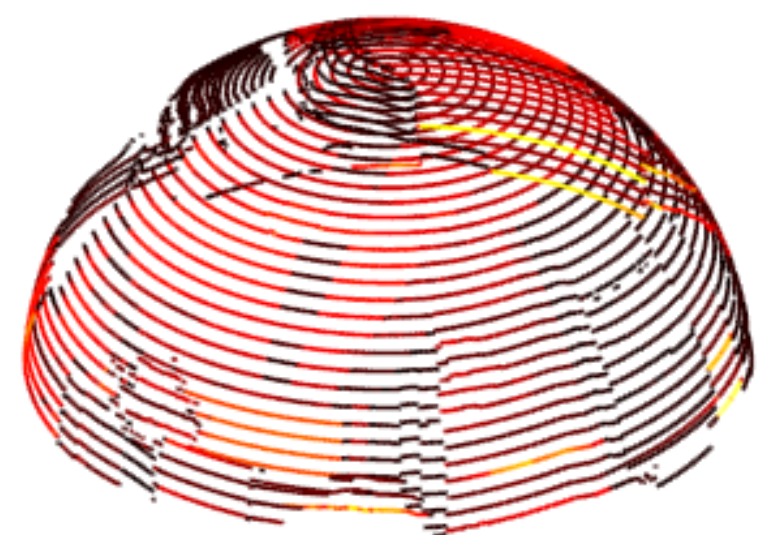}
        \label{fig:bpearl}
    }\hfill
    \subfloat[OS0-128] {%
        \includegraphics[width=0.27\columnwidth, trim={6.05cm 2.0cm 5.85cm 0.0cm}, clip]{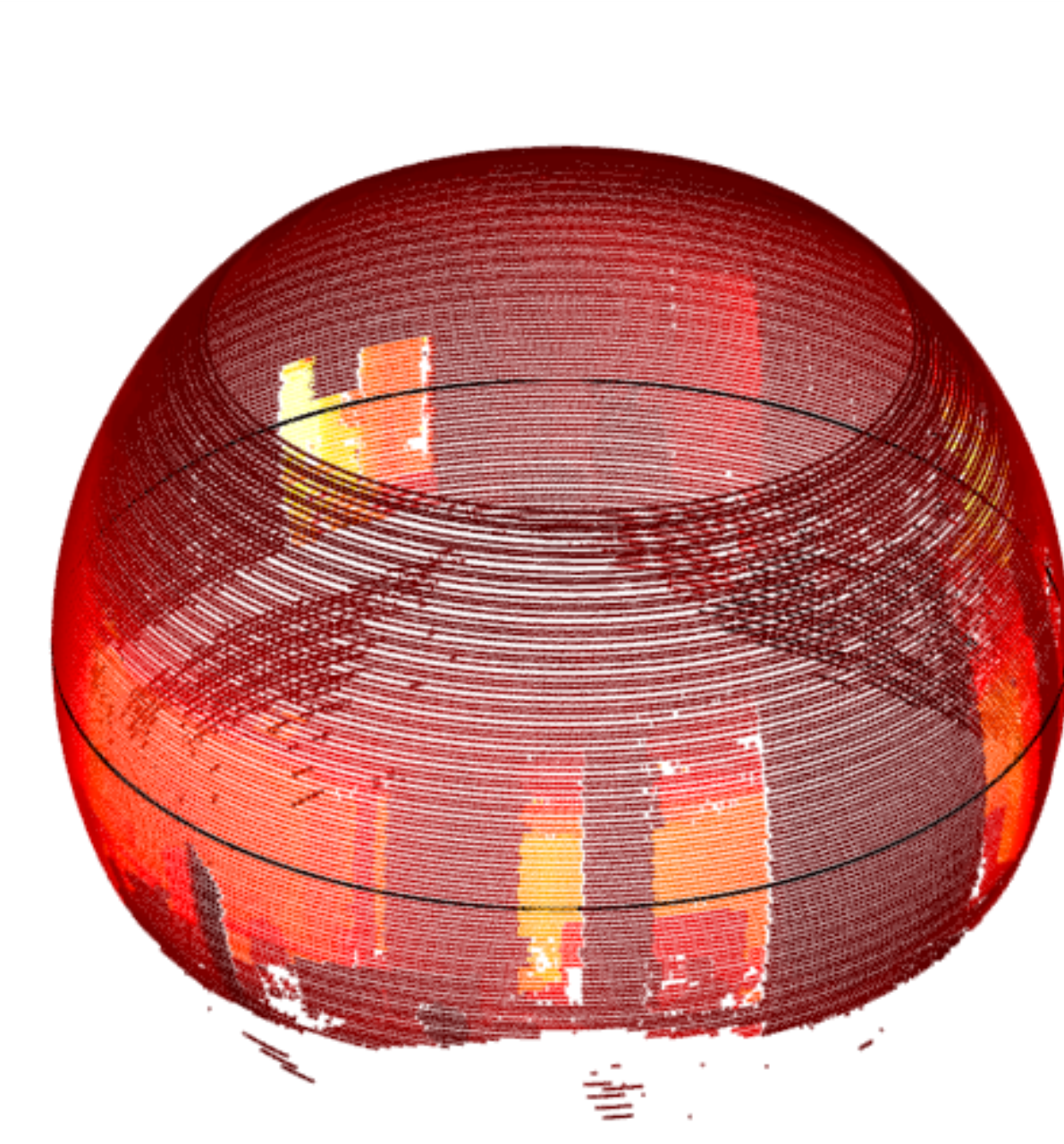}
        \label{fig:os0}
    } 
    \caption{Illustration of the resulting spherical projections using different LiDAR sensors. Projection (a) shows the result of an Ouster OS1 with 64 beams, (b) a Robosense-BPearl with 32 beams, and (c) an Ouster OS0 with 128 beams.}
    \label{pics:projections}
    \vspace{-5mm}
\end{figure}
As an initial step, we project each pointcloud onto the sphere, i.e. the $j$th point $\vec{p}_j=[x_j,y_j,z_j]^\top$ is projected using
\begin{align}
    \phi_j=\arctan\left(\frac{y_j}{x_j}\right),\hspace{1cm}
    \theta_j=\arccos\left(\frac{z_j}{||\vec{p}_j||_2}\right),
\end{align}
where $\phi_j$ is the azimutal and $\theta_j$ the polar angle as defined in Section~\ref{sec:pre:fourier}.
Figure~\ref{pics:projections} illustrates the projections of three different LiDAR sensors. 

After the projection, we employ the sampling theorem by Driscoll and Healy~\cite{Driscoll1994} to create a $2\tilde{B}\times2\tilde{B}$ sampling grid on the sphere.
We employ a nearest neighbor search using the grid points to get the sampled pointcloud where we average points that fall within the same bin, s.t. this discretization yields $\tilde{f},\tilde{h}\in\mathbb{R}^{2\tilde{B}\times2\tilde{B}\times{}K}$. 
This enables the use of the spherical Fourier transform (cf. eq.~\eqref{eq:pre:spherical_fourier}) yielding $\tilde{F}$ and $\tilde{H}$. 
The spherical correlation $\tilde{\mat{C}}$ of two functions $\tilde{f}$ and $\tilde{h}$ is defined as their inner product, i.e. $\tilde{\mat{C}} = \langle\tilde{f},\tilde{h}\rangle$. 

If $\tilde{h}$ is a rotated version of $\tilde{f}$, we find the rotation $\vec{r}\in SO(3)$ by maximizing $\tilde{\mat{C}}$, i.e.
\begin{equation}\label{eq:method:spherical_correlation}
     \underset{\vec{r}\in{}SO(3)}{\mathrm{arg \, max}} \langle\tilde{f}, \vec{r}^{-1}\tilde{h}\rangle,
\end{equation}
where the inner product is defined as in eq.~\eqref{eq:pre:s2_integral}.
As proposed by Kostelec and Rockmore~\cite{Kostelec2008}, by making use of eq.~\eqref{eq:pre:fourier_expansion} and the fact that spherical harmonics are orthogonal, eq.~\eqref{eq:method:spherical_correlation} can be rewritten and efficiently evaluated using the spherical Fourier coefficients, i.e.
\begin{equation} \label{eq:method:spherical_correlation_fourier_coefficients}
    \tilde{\mat{C}} = \sum_{l>0}\sum_{m\le l}\sum_{m'\le l} \tilde{F}_m^l \overline{\tilde{H}_{m'}^l D_{mm'}^l},
\end{equation}
where $D$ denotes the Wigner-D functions.
For the reader interested in a detailed derivation of eq.~\eqref{eq:method:spherical_correlation_fourier_coefficients} please refer to~\cite{Kostelec2008}.

\begin{figure}[!thb]
\vspace{-3mm}
\centering
    \subfloat[Quaternion samples] {%
        \includegraphics[width=0.29\columnwidth, trim={11.5cm 7.0cm 11.5cm 7.3cm}, clip]{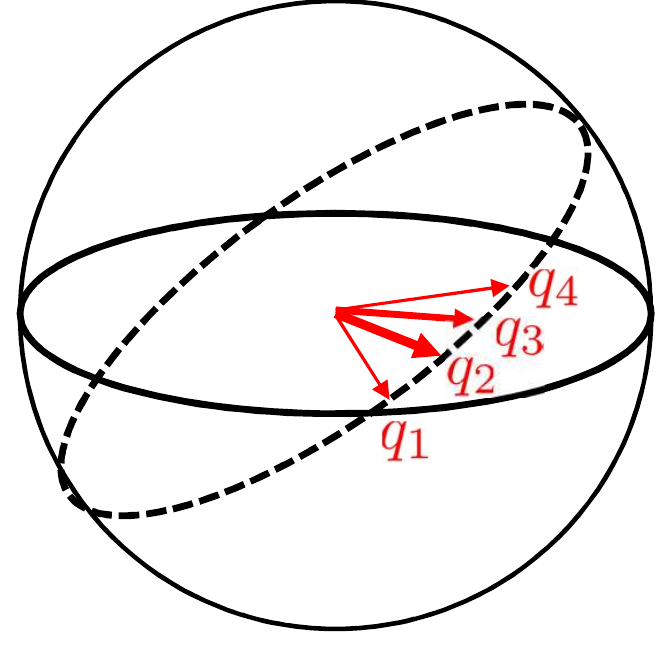}%
        \label{fig:quaternion}
    } \hspace{1cm}
    \subfloat[Bingham distribution] {%
        \includegraphics[width=0.29\columnwidth, trim={6cm 9cm 5.2cm 8.6cm}, clip]{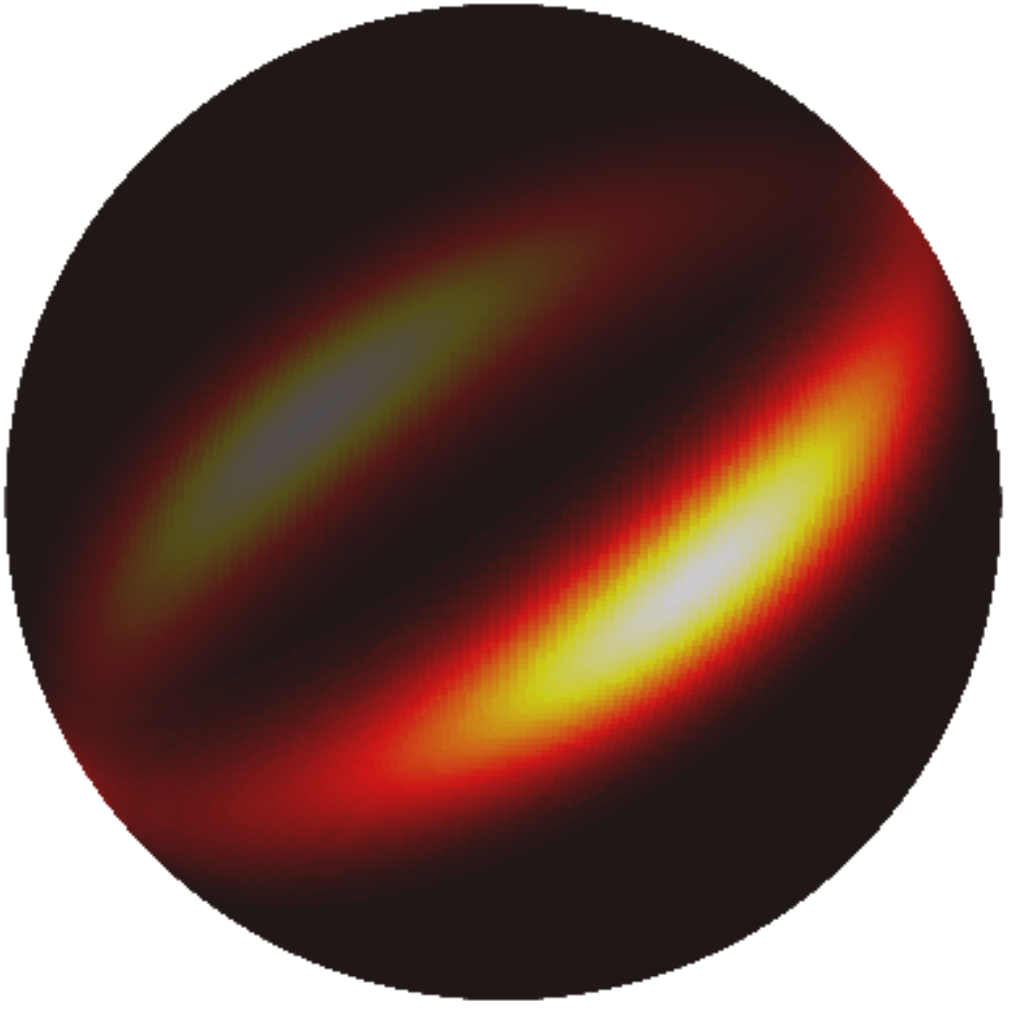}
        \label{fig:bingham}
    }
    \caption{(a) Illustrates different rotational estimates on the unit sphere. The thickness of the arrows denotes the weight of the estimate. (b) Shows a fitted Bingham distribution from weighted samples. Generally, noise and partial overlap result in high weighted candidates and can perturb correlations. Our proposed fusion of multiple modalities alleviates the correlation bias and yields a more precise result.}
    \label{pics:bingham:fitted}
    \vspace{-3mm}
\end{figure}

After correlating the coefficients, we zero-pad the coefficients vector to increase the resulting resolution and perform a subsequent inverse $SO(3)$ Fourier transform~\cite{Kostelec2008} to find the rotation $\vec{r}$ that maximizes eq.~\eqref{eq:method:spherical_correlation}. Thereby, we find the maximum, which corresponds to our best estimate for the rotational alignment and its neighbors in the correlation tensor.
In concrete, each signal in the correlation tensor is a rotation that is initially represented as ZYZ-Euler angles. 
Using the normalized magnitude as the weight, we convert the ZYZ-Euler angles to a unit-quaternion and fit the maximum peak as well as its neighbors to a Bingham distribution (cf. Figure~\ref{pics:bingham:fitted}). 
In more detail, we perform a maximum likelihood estimation by matching the second-order moment with the unit-quaternion samples $\mat{Q}_\mathcal{B}$ and their corresponding weights $\vec{w}_\mathcal{B}$, i.e.
\begin{equation}
    \Sigma_\mathcal{B} = \mat{Q}_\mathcal{B}\mat{W}_\mathcal{B}\mat{Q}_\mathcal{B}^\top = \mat{U}\Lambda\mat{U}^\top,
\end{equation}
where $\mat{W}_\mathcal{B}=\operatorname{diag}(\vec{w}_\mathcal{B})$, and where $\mat{\Lambda}=\operatorname{diag}(\vec{\lambda})$ and $\mat{U}$ are the eigenvalues and eigenvectors after an eigendecomposition of $\Sigma_\mathcal{B}$. 
Using the relationship in eq.~\eqref{eq:simga_b}, the eigenvectors $\mat{U}$ correspond to the orientation $M$, and $\lambda_i$ corresponds to $\frac{\frac{\partial}{\partial z_i}N(\mat{Z})}{N(\mat{Z})}$. 
Finally, we use a local and convex optimization to derive the optimal concentration parameter $\hat{\mat{Z}}$ using the following cost function
\begin{equation}
    J(\hat{\mat{Z}}) = \sqrt{\sum_i\left(\frac{\frac{\partial}{\partial \hat{z}_i}N(\hat{\mat{Z}})}{N(\hat{\mat{Z}})} - \vec{\lambda}_i\right)^2},
\end{equation}
where $\vec{\lambda}_i$ is the $i$-th normalized eigenvalue. 
The fitted Bingham distribution comprises the rotational estimation part of the registration.
A limitation of our uncertainty derivation is that this does not capture the uncertainty of other potential peaks in the correlation but comprises only the highest correlating rotation. 

At this point, we take one of the modes of Bingham and rotate the source pointcloud accordingly. 
Since the Bingham distribution is antipodal symmetric, it is indistinctive which mode we take.

\subsection{Translation Estimation}\label{sec:method:tran}
After estimating the rotation, the source pointcloud is correctly oriented, enabling us to estimate the translation using a correlation for the full alignment. 
We first discretize the pointcloud space $\mathbb{R}^3$ in $n$ equal dimensions with zero-padding to satisfy the requirements of eq.~\eqref{eq:pre:spatial_fourier}. 
The two discretized pointclouds $f, h\in\mathbb{R}^{n\times{}n\times{}n\times{}K}$ from Section~\ref{sec:method:representation} are transformed to the spatial frequency domain using eq.~\eqref{eq:pre:spatial_fourier}, resulting in $F$ and $H$, respectively.
Similar to the rotational registration, we perform another correlation $C$ of the coefficients to estimate the phase shift, i.e.
\begin{equation}
    C = F\cdot{}\overline{H},
\end{equation}
where $\overline{H}$ is the complex conjugate of $H$.
Zero-padding as well as a subsequent inverse Fourier transform of $C$, 
\begin{align}
c(x,y,z) = \sum_{u=0}^{N-1}&\sum_{v=0}^{N-1}\sum_{w=0}^{N-1} C(u,v,w) \nonumber\\
  &\exp\left(j\frac{2\pi}{N} [x,y,z]^\top[u,v,w]\right),
\end{align}
yields the final translation by finding the most prominent peaks of $c(\vec{x})$. 
In a similar fashion as in Section~\ref{sec:method:rot}, we search for the maxima in the correlation function. 
Again, each signal defines a specific relative translation weighted by the magnitude, and the maximum corresponds to the best estimate of it. 
Using the maximum and its neighbors as samples $\mat{T}$ we fit a multivariate Gaussian distribution where the mean $\vec{\mu}$ is the weighted average and covariance $\Sigma_\mathcal{N}$ is defined as
\begin{equation}
    \Sigma_\mathcal{N} = (\mat{T} - \vec{\mu}) \mat{W}_\mathcal{N} (\mat{T} - \vec{\mu})^\top,
\end{equation}
where $\mat{W}_\mathcal{N} = \operatorname{diag}(\vec{w}_\mathcal{N})$ is a diagonal matrix of individual weights $\vec{w}_\mathcal{N}$.

This results in our translational estimate, and therefore we have derived the relative transformation between the source and the target pointcloud for a single modality.
\subsection{Multi-Modal Fusion}\label{sec:method:fusion}
The previous sections cover the registration core of PHASER using a single modality. 
Given a set of representative functions (cf. Section~\ref{sec:method:representation}) transformed in the spherical and spatial frequency domain, we now seek to find a fused spectrum for source and target pointcloud. 
Having a single fused spectrum allows us to leave the correlation pipeline, i.e., registration core of PHASER, to remain unchanged. 

Spectral image fusion techniques~\cite{Naidu2011} combine multiple complementary sensors to create an enhanced image by analyzing the spectral bands and selecting the most relevant parts of each sensor~\cite{Falk1999}. 
Inspired by these techniques, we employ a Laplacian pyramid-based fusion to construct a fused pointcloud spectrum from different modalities. 
 
Initially, we construct a Gaussian pyramid where each level represents a low-pass filtered and subsampled by half version of the previous level. 
The difference of each level to the previous level denotes the Laplacian pyramid.
At the first level, we utilize the difference to the original pointcloud spectrum.
Range and intensity information yield a good spectral diversity, and when fused, provide a more enriched representation of the environment. 
Consequently, we construct a Laplacian pyramid for each input channel representing the details (contrast changes) for each channel. 
We fuse the channels by reconstructing a single merged pointcloud spectrum by combining the Laplacian pyramids to a single pyramid. 
Specifically, the top-level low-pass spectra are averaged, whereas all the other levels select the coefficient with the highest local energy~\cite{Falk1999}. 
Finally, the fused spectrum will be the input for the registration pipeline, as discussed in Section~\ref{sec:method:rot} and Section~\ref{sec:method:tran}.
\newcommand{\PrecisionTab}{
    \centering
    \begin{tiny}
    \begin{tabular}[b]{ *4c }
    \toprule
    PSNR & 50 & 45 & 40 \\
    \midrule     
    ICP & 100\% & 100\% & 100\%  \\
    ICP-Plane & 100\% & 96\% & 96\% \\
    GICP & 100\%  & 100\% & 88\% \\
    NDT & 100\% & 100\% & 100\% \\
    CPD & 100\% & 100\% & 100\% \\
    \midrule
    Super4PCS & 60\% & 57\% & 53\% \\
    Go-ICP & 42\% & 42\% & 42\% \\
    SDRSAC & 77\% & 73\% & 80\% \\
    FastDesp & 48\% & 41\% & 33\% \\
    Teaser++ & 100\% & 100\% & 100\% \\
    PH-Range & 100\% & 100\% & 100\% \\
    PH-Fused & 100\% & 100\% & 100\% \\
    \bottomrule
    \end{tabular}
    \end{tiny}
}
\begin{figure*}[!htb]
  \centering
   \includegraphics[width=1\textwidth, trim={0.0cm, 0.0cm, 0.0cm, 0cm}, clip]{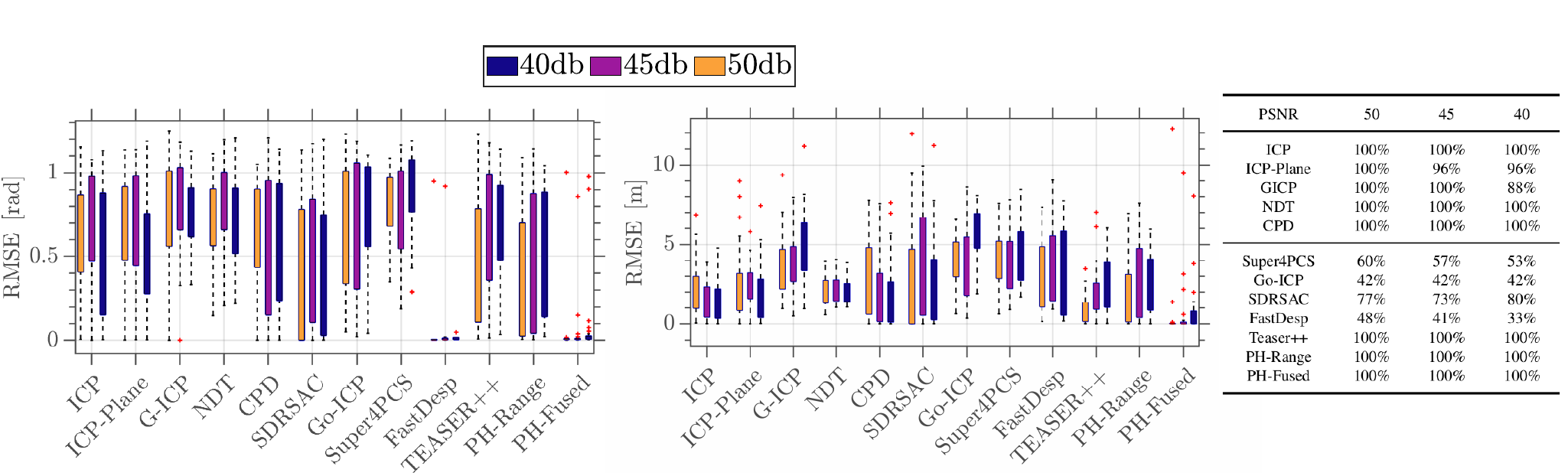}
   \vspace{-9mm}
   \caption{Evaluation of the RMSE in terms of rotation (left) and translation (middle) at different PSNR levels. We perform 80 random registrations for each PSNR level (240 in total) and show each method's success rate in the table (right). Here, \textit{ICP-Plane} refers to the point-to-plane approach. \textit{PH-Range} and \textit{PH-Fused} denote PHASER using range-information only and a fusion of the available channels, respectively.}
   \label{pics:exp:simulated_rmse}
   \vspace{-5mm}
\end{figure*}
\section{Experiments}
\label{sec:experiments}
This section presents the experimental validation of our global alignment framework, where we show that our pipeline is robust against noise as well as partial overlap, converges to the global solution, and provides a meaningful uncertainty estimate.
Throughout this section, we consider four local (ICP~\cite{Besl1992}, NDT~\cite{Magnusson2007}, CPD~\cite{Myronenko2010}, G-ICP~\cite{Segal2009}) and five global (SDRSAC~\cite{Le2019}, FastDesp~\cite{Lei2017}, Go-ICP~\cite{Jiaolong2015}, Super4PCS~\cite{Mellado2014}, TEASER++~\cite{Yang2020}) registration methods for comparison.
We used the respective authors' implementation except for ICP, NDT and CPD where we utilized the implementation that is available in MATLAB.
For TEASER++, we first computed correspondences using FPFH~\cite{Rusu2009} before estimating the relative transformations.
We configure PHASER with $\tilde{B}=120$, $B=6$, i.e. $\tilde{f},\tilde{h}\in\mathbb{R}^{240\times240\times{}K}$ and $f,h\in\mathbb{R}^{200\times200\times200\times{}K}$.
Moreover, we use five pyramid levels and four neighbors for fitting the Bingham and Gaussian distribution. 
All experiments have been performed on real data and include a mix of indoor and outdoor environments.
\subsection{Synthetic Benchmarks}\label{sec:exp:sim_gt}
%
In this experiment, we collected 60 pointclouds from robots equipped with a 64 beam Ouster OS-1 (65535 points per cloud) and 20 pointclouds from the KITTI dataset~\cite{Geiger2012} (Sequence 00) to simulate a global alignment problem. 
From the OS-1 dataset, we use four channels ($K=4$): range, intensity, reflectivity, and ambient IR, whereas from the KITTI dataset, we utilize range and intensity ($K=2$).

\textbf{Noisy Measurements.} We applied a random transformation to each pointcloud as well as additive zero-mean Gaussian noise to all of its channels. 
This results in a bijection between the transformed source and the initial target pointcloud. 
\begin{figure}[!t]
  \centering
   \includegraphics[width=0.5\textwidth, trim={0.0cm, 0cm, 0.0cm, 0cm}, clip]{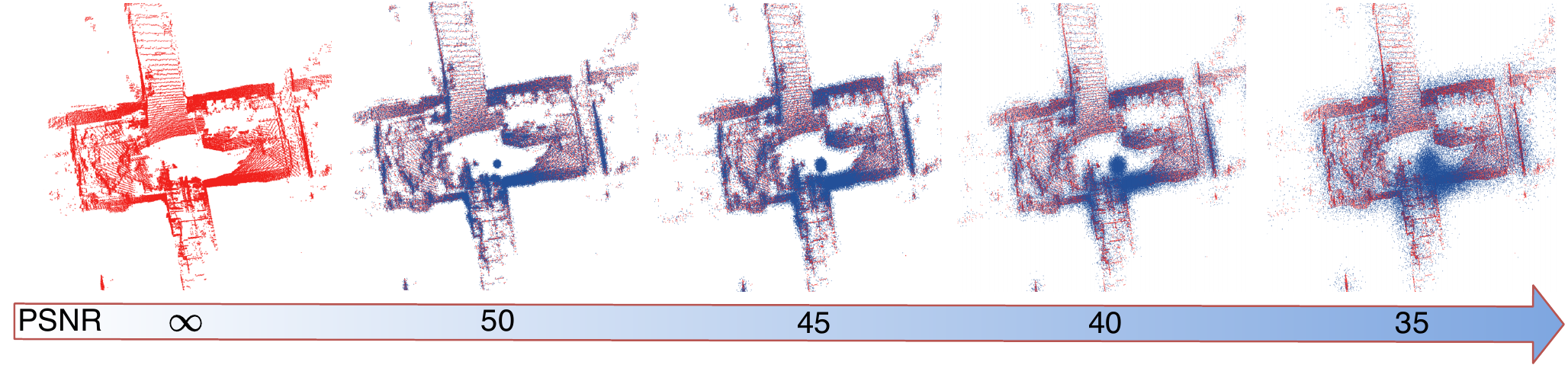}
   \vspace{-8mm}
   \caption{Illustration of different PSNR levels. Red denotes the original pointcloud, blue the noised one. Our proposed pipeline is capable of robustly aligning pointclouds up to $40\,\mathrm{dB}$ PSNR.}
   \label{pics:exp:psnr}
   \vspace{-6mm}
\end{figure}
In particular, the experiments presented in this section aim to validate that our pipeline converges to the global solution independent of the initial alignment, i.e., it serves as a global registration method.
Figure~\ref{pics:exp:simulated_rmse} shows the evaluation of the simulated transformations given different noise levels (cf.~Figure~\ref{pics:exp:psnr}) as well as their success rate. 
We consider an estimation unsuccessful if either the program crashes or the registration takes longer than 30 minutes.

Due to many local minima and limited search spaces, all local approaches cannot register the test cases correctly.
The global search of Go-ICP provides a more extensive search space, but its least square objective is erroneous under the presence of noise.
Super4PCS fails for all cases to correctly estimate the transformation since the noise disables the congruence search in the target pointcloud. 
SDRSAC showed better robustness due to its randomized approach yielding a broad error distribution.
FastDesp and TEASER++, show competitive results despite being highly influenced by the noise during the descriptor extraction.
Although our method is correspondence-free, it is similarly affected by the noise if we only consider the range information (\textit{PH-Range}).
However, if we supply our pipeline with all channels, we show that the registration is greatly improved (\textit{PH-Fused}) even in high noise cases ($40\,\mathrm{dB}$) where we achieve an improvement of $91\,\%$ and $73\,\%$ in terms of the rotational and translation RMSE, respectively. 
Additionally, we achieve a similar rotational RMSE compared to FastDesp but gain a better success rate and an improvement of $80\,\%$ for the $40\,\mathrm{dB}$ translational estimation. 
For TEASER++, we yield a $93\,\%$ and $80\,\%$ improvement w.r.t. to the $40\,\mathrm{dB}$ rotational and translational RMSE, respectively.

\begin{figure}[!htb]
\vspace{-3mm}
\captionsetup[subfigure]{labelformat=empty}
\centering
    \includegraphics[width=0.51\textwidth, trim={0.37cm, 3.0cm, 0.4cm, 0.065cm}, clip]{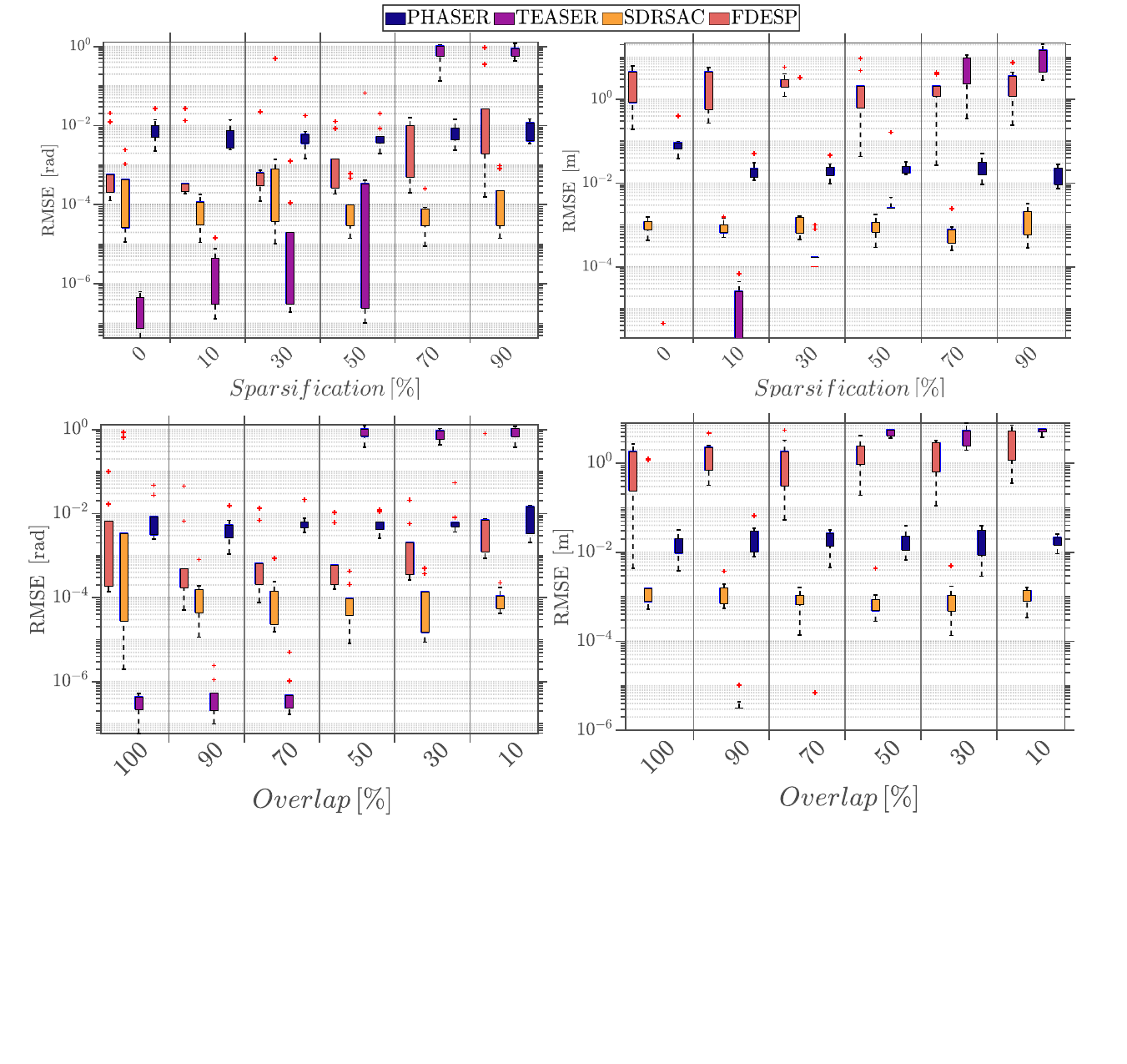}%
    \label{pics:exp_overalp:trans}
    \vspace{-6mm}
 \caption{Distribution of the RMSE of rotation (left) and translation (right) w.r.t. simulated sparsification (top) and simulated overlap (bottom) in log scale. Results are shown over 30 pointclouds registrations.}
 \label{pics:exp_overlap2}
 \vspace{-3.5mm}
\end{figure}

\textbf{Missing Data.} Next, we remove the artificial noise from the experiments and investigate the low overlapping case.
Removing the noise does not corrupt the descriptor generation anymore, but instead, with little overlap, we provide less information to it.
Consequently, we propose two experimental setups. 
First, we sparsified the pointclouds by randomly removing points from the source cloud before applying a random transformation. 
Second, given a percentage of points, we removed slices starting from the maximum towards the minimum resulting in pointclouds that only partially observed an environment, i.e., a simulated partial overlap.
Here, we further investigate the performance of SDRSAC, TEASER++ and FastDesp compared to PHASER since those yielded competitive results in the first experiment.
Figure~\ref{pics:exp_overlap2} illustrates the evaluation given different sparsification rates and overlap (partial observability) in log scale.
In this experiment, SDRSAC attains significantly better registration quality since its graph matching performs better on noiseless data.
For TEASER++, the RMSE starts low with little sparsification and high overlap and increases with large sparsification and little overlap.
Removing large portions of the pointclouds leads to less precise descriptors for the source pointcloud resulting in little inlier rates.
The multiscale descriptor approach of FastDesp yielded good rotational registration but failed to infer the translation for all of our datasets accurately.
PHASER consistently estimates the transformation even at $90\%$ sparsification and $10\%$ overlap and maintains an RMSE of $0.4^\circ$ and $0.15\,\mathrm{m}$ as we do not rely on any correspondences but rather measure the similarity of source and target pointcloud.
Concluding, our approach enables accurate global registrations with noise and partial overlap.
Our precision directly links to the spherical and spatial bandwidth and its resulting discretization, which we discuss in the next section. 
\subsection{Performance Evaluation}\label{sec:exp:performance}
\begin{figure}[!htb]
\vspace{-4mm}
\captionsetup[subfigure]{labelformat=empty}
\centering
    \subfloat[] {%
        \includegraphics[width=0.24\textwidth, trim={0.3cm, 0.2cm, 0.4cm, 0.6cm}, clip]{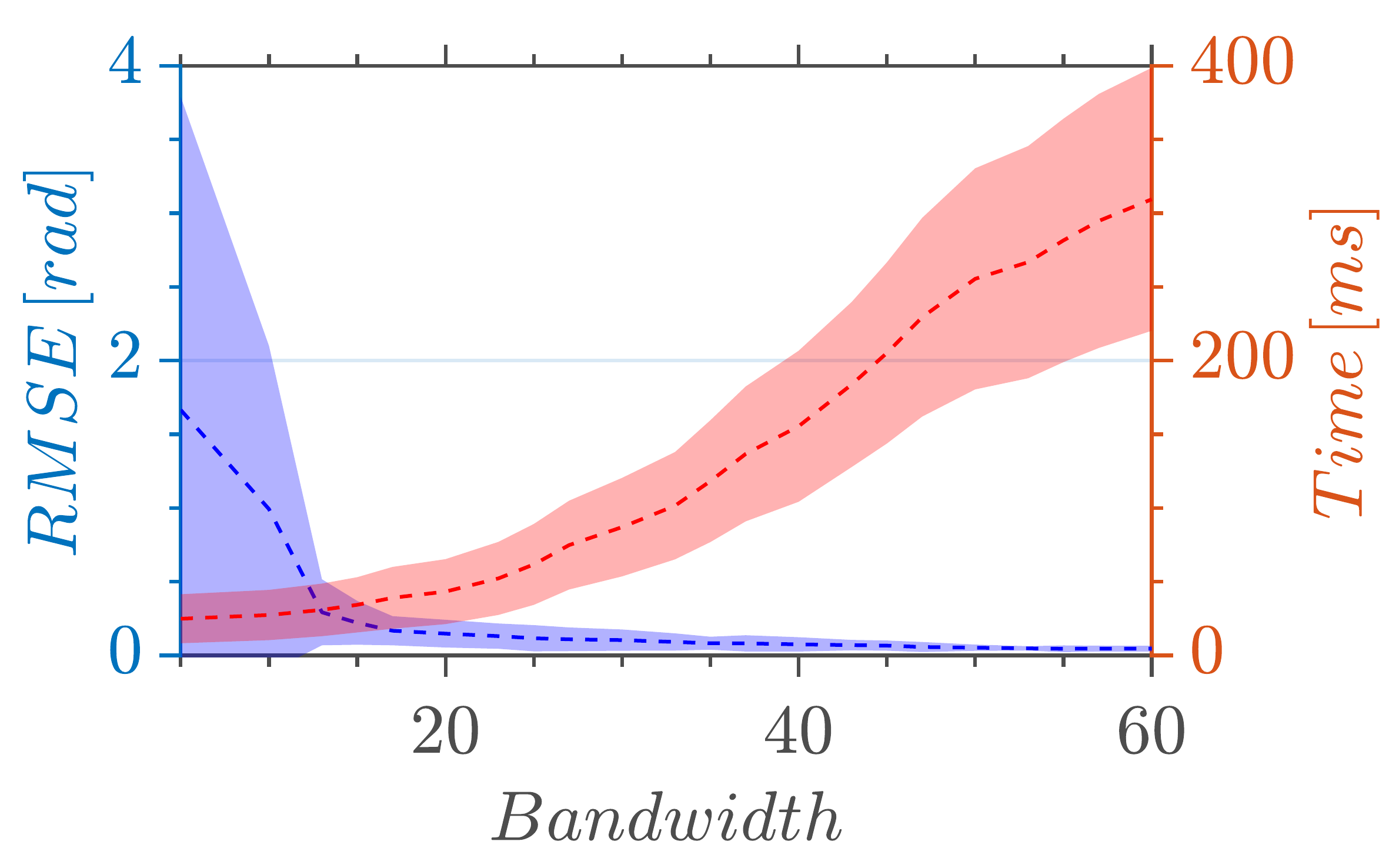}%
        \label{pics:exp2:execution_r}
    }
    \subfloat[] {
        \includegraphics[width=0.24\textwidth, trim={0.45cm, 0.2cm, 0.2cm, 0.6cm}, clip]{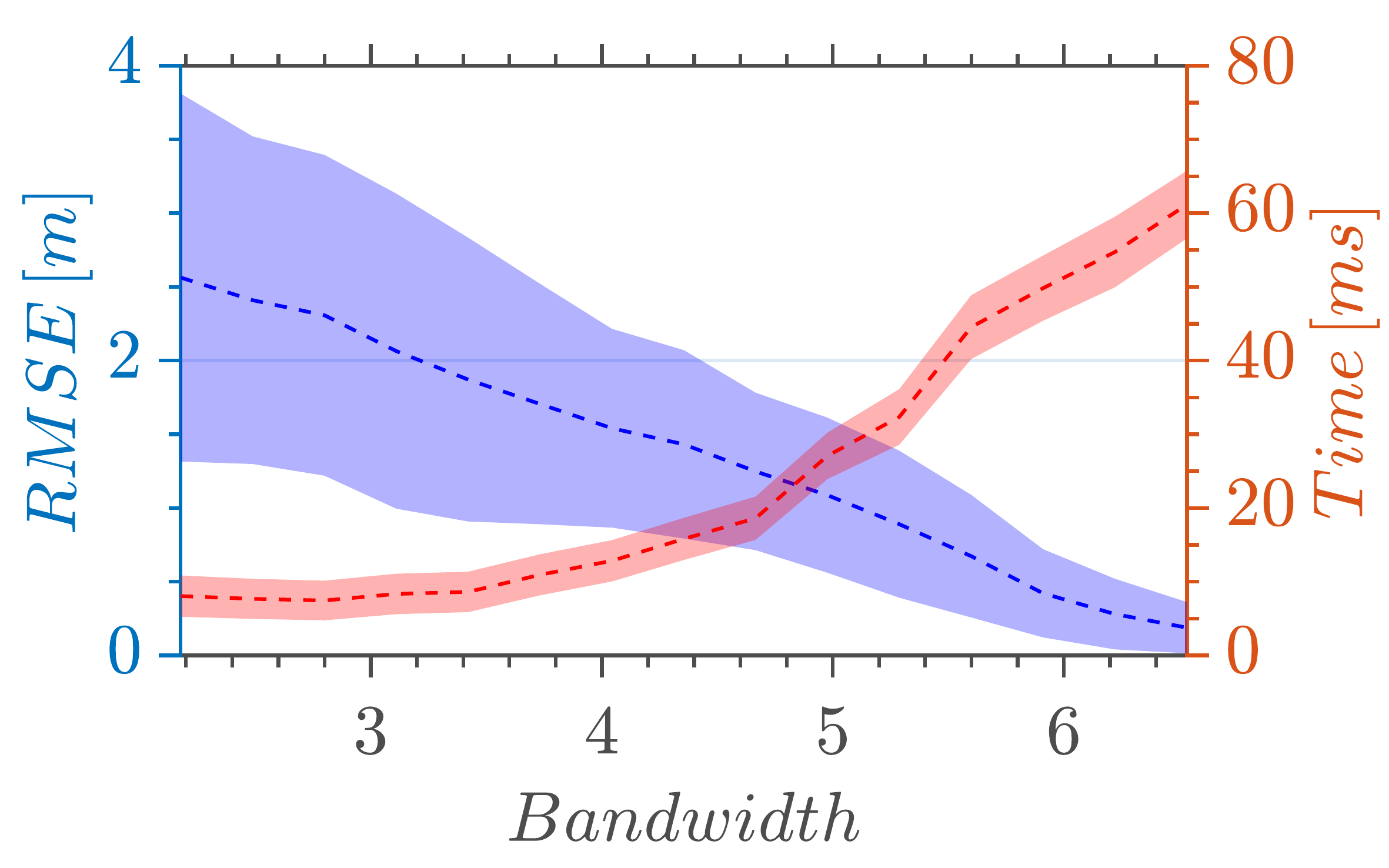}%
        \label{pics:exp2:execution_t}
    }
    \vspace{-6mm}
 \caption{Rotational (left) and translational (right) error w.r.t. computation time for a single modality. The execution time incorporates the discretization on $S^2$ and $\mathbb{R}^3$ as well as the individual correlations.}
 \label{pics:exp2:execution}
 \vspace{-6mm}
\end{figure}

PHASER contains two main parameters, namely the spherical ($\tilde{B}$) and spatial ($B$) bandwidth. 
Both parameters control the registration pipeline, e.g., to have a sparse or dense sampling of the space. 
Figure~\ref{pics:exp2:execution} shows the trade-off between execution time and alignment error using different bandwidths. 
All experiments were executed on a laptop computer with an Intel i7-8850H.
To summarize, our global alignment algorithm can be tailored to a coarse or fine registration by varying $\tilde{B}$ and $B$. 
Our experiments show that a small spherical bandwidth (20-30) already yields a reasonable rotational estimate and, together with a spatial bandwidth of 6-7 (cf. Figure~\ref{pics:exp2:execution}), provides a good global alignment in less than $100\,\mathrm{ms}$. 
\subsection{Uncertainty Analysis}
\begin{figure}[!htb]
\vspace{-6mm}
\captionsetup[subfigure]{labelformat=empty}
\centering
    \subfloat[] {%
        \includegraphics[width=0.243\textwidth, trim={0.1cm, 0.0cm, 0.8cm, 0.7cm}, clip]{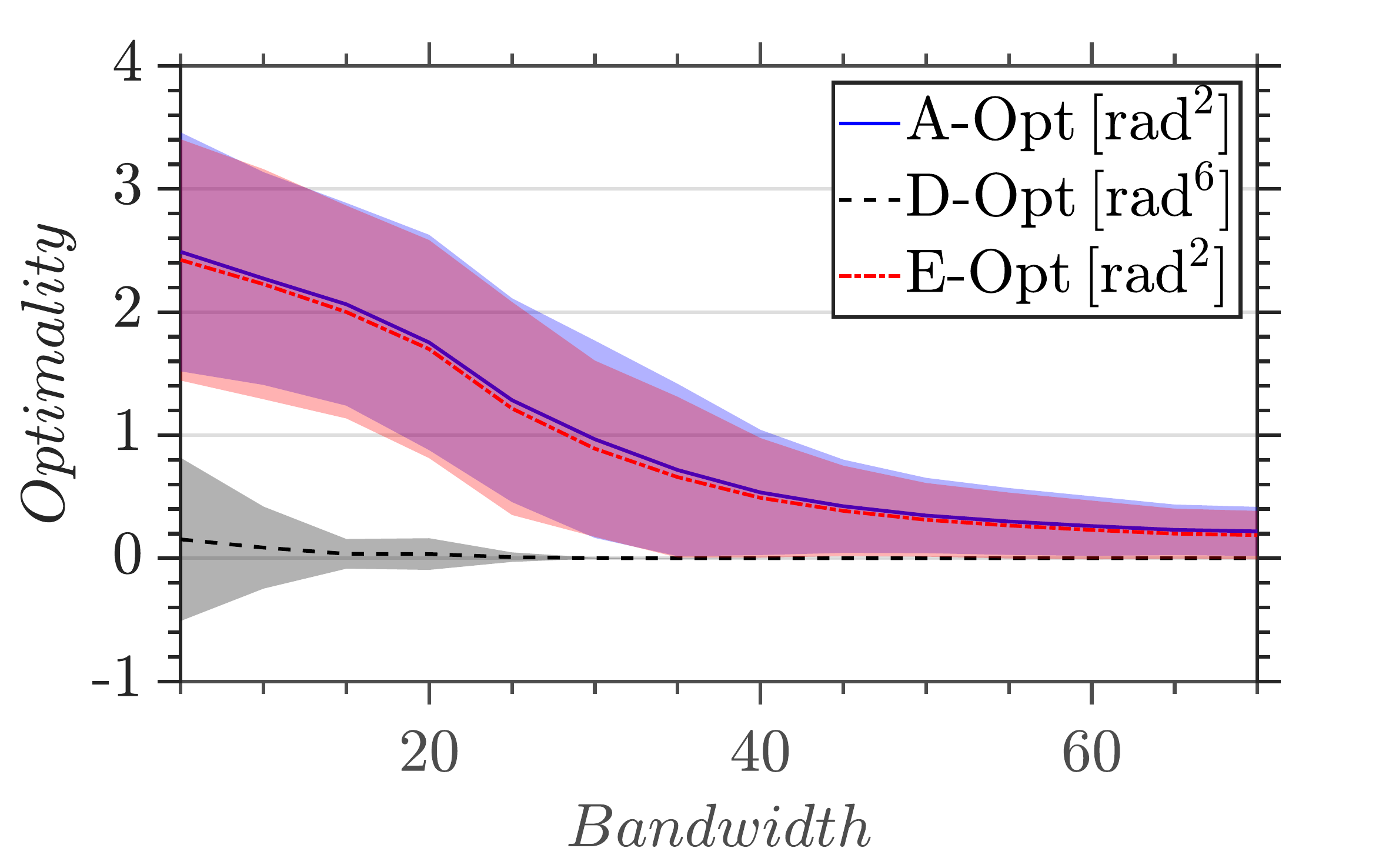}%
        \label{pics:exp_overalp:rot}
    }
    \subfloat[] {%
        \includegraphics[width=0.243\textwidth, trim={0.1cm, 0.0cm, 0.8cm, 0.7cm}, clip]{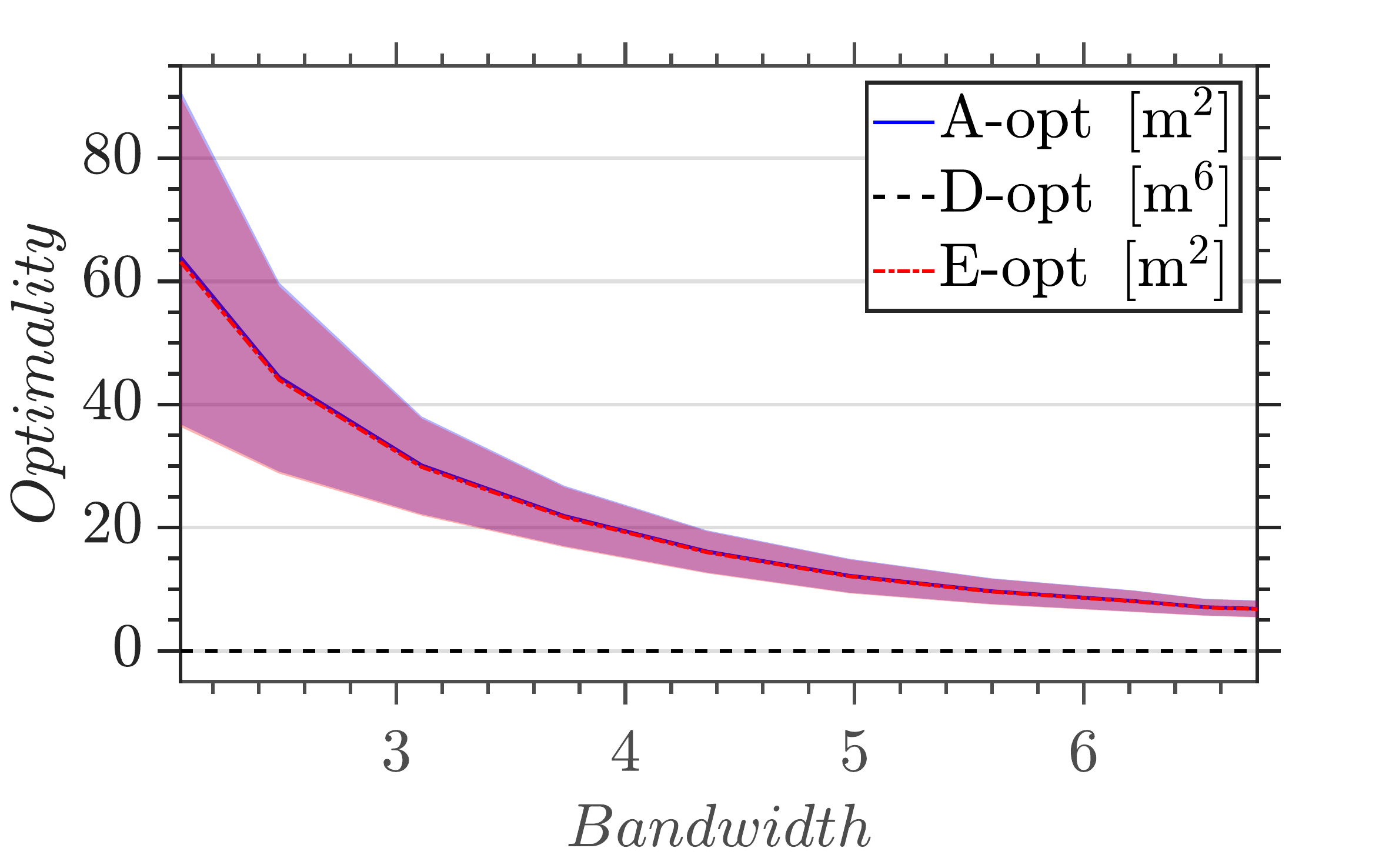}%
        \label{pics:exp_overalp:trans}
    }
    \vspace{-6mm}
    \caption{Evaluation of the rotational covariance $\Sigma_{\alpha\beta\gamma}$ (left) and the positional covariance $\Sigma_\mathcal{N}$ (right) for different bandwidths. The line corresponds to the mean, and the shaded area denotes the standard deviation for each criterion.}
    \label{pics:uncert}                                             
    \vspace{-3mm}
\end{figure}

In this section, we evaluate our uncertainty estimate based on three criteria~\cite{Carrillo2012}: (i) the trace of the covariance (A-opt), (ii) the determinant of the covariance (D-opt), and (iii) the maximum eigenvalue of the covariance (E-opt). 
Figure~\ref{pics:uncert} shows the evaluation of these criteria, based on the dataset used in Section~\ref{sec:exp:sim_gt} with $PSNR=50\,\mathrm{dB}$.
These results infer that the uncertainty is consistent with the registration error (cf. Figure~\ref{pics:exp2:execution}), i.e., it decreases with increasing bandwidth and hence with decreasing error. 

Note that for better visualization of the rotational uncertainty, we created a sample-based Gaussian covariance of the half-sphere and converted it to an Euler angle covariance, $\Sigma_{\alpha\beta\gamma}$, using the covariance laws defined by Vanicek and Krakiwsky~\cite{vanicek2015geodesy}.

\subsection{Real World Experiments}
\begin{figure}[!htb]
\vspace{-6mm}
\captionsetup[subfigure]{labelformat=empty}
\centering
    \subfloat[] {%
        \includegraphics[width=0.248\textwidth, trim={0.1cm, 0.13cm, 0.3cm, 0.9cm}, clip]{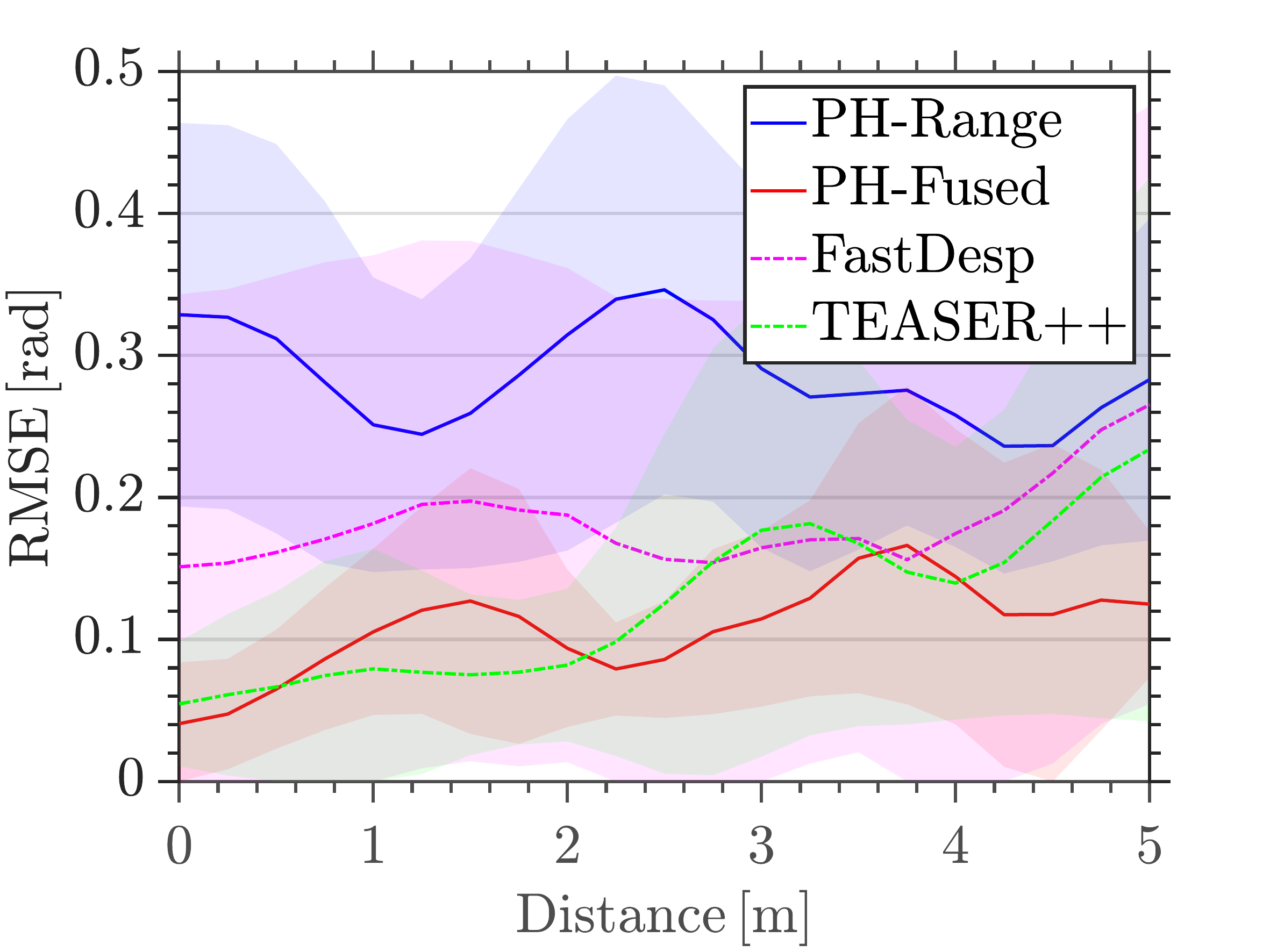}%
        \label{pics:exp:rw_rot_improvement}
    }
    \subfloat[] {%
        \includegraphics[width=0.248\textwidth, trim={0.2cm, 0.13cm, 0.3cm, 0.9cm}, clip]{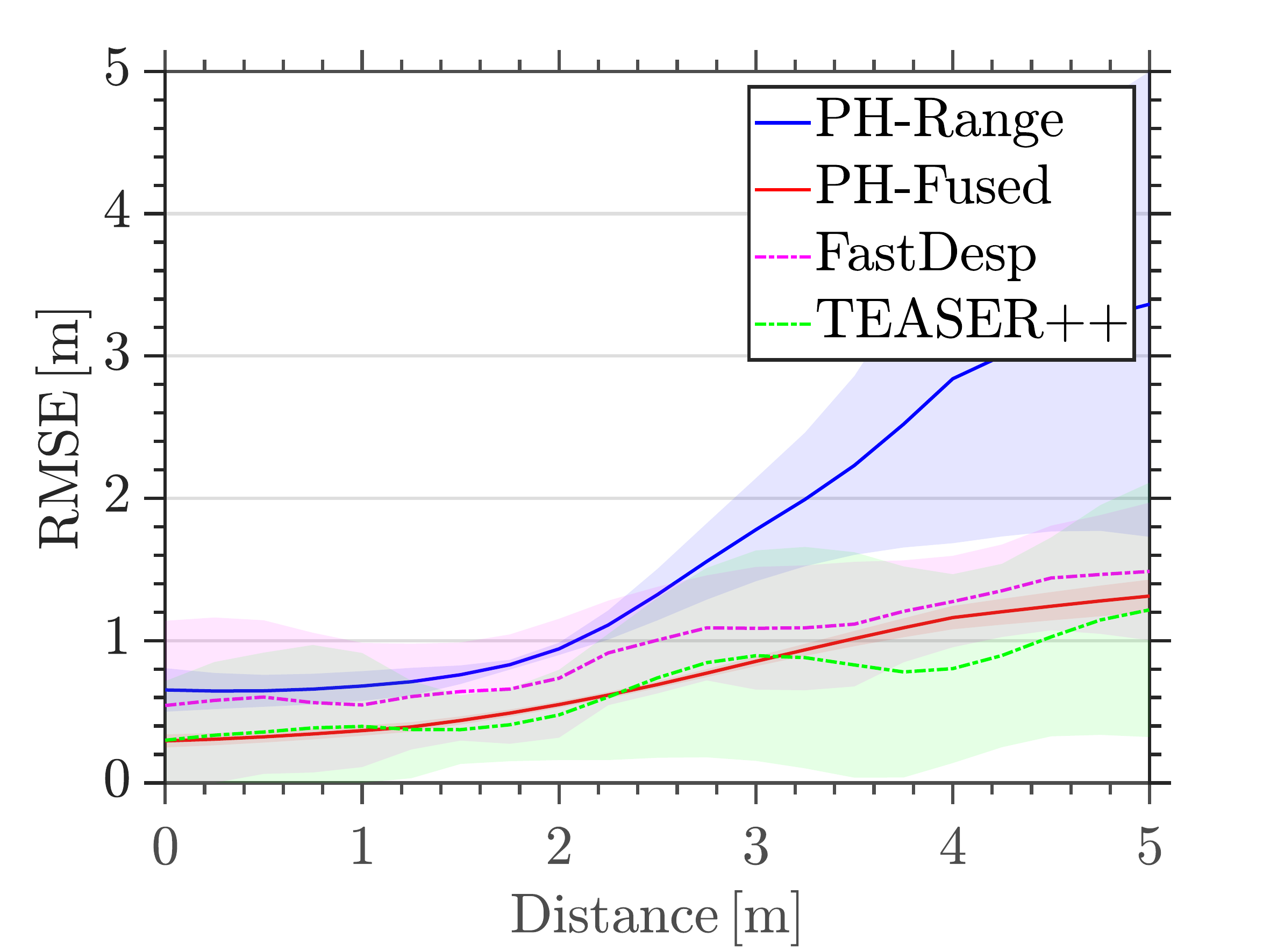}%
        \label{pics:exp:rw_trans_improvement}
    }
    \vspace{-5mm}
 \caption{Rotational (left) and translational (right) error w.r.t. distance between the poses. Utilizing a fusion of the range values with the intensity channel allows our pipeline to automatically choose the better modality for the registration and decrease the error accordingly.}
 \label{pics:exp:rw_improvement}
 \vspace{-6.5mm}
\end{figure}

We recorded several runs in a search and rescue testing facility using a handheld device with an RTK GPS and a 128 beam Ouster OS-0 (131072 points per scan).
The environment comprises urban-like streets, buildings, and collapsed structures. 
In this experiment, we only utilize the range and intensity channels ($K=2$).
Given the GPS map ($10\,\mathrm{km}$ long) with poses and scans, we randomly sampled inter and intra run pairs based on their positional proximity to evaluate how the distance affects our registration pipeline. 
As ground truth, we considered the relative GPS poses refined by \ac{ICP} and grouped them by their relative distance. 
Specifically, in an interval of $0.25\,\mathrm{m}$, we averaged over 20 registration each (420 registrations in total). 
%
Figure~\ref{pics:exp:rw_improvement} illustrates the improvement of our proposed Laplacian-based fusion compared to a registration using range information only, TEASER++ and FastDesp.
As expected, the registration using range information achieves a broad error distribution over the different distances while our proposed fusion stays mostly concise with an average RMSE of $6^\circ$ and $0.7\,\mathrm{m}$. 
Furthermore, we show that especially at greater distances ($4\text{-}5\,\mathrm{m}$) TEASER++ and FastDesp have increased rotational error due to bad correspondences, whereas PHASER still maintains a low error.
\section{Conclusion and Future Work}
\label{sec:summary}
We have presented PHASER, a new framework for global pointcloud alignment based on the correlation of Fourier transformed signals. 
We make use of translational and rotational invariance of the spatial and spherical Fourier transform to estimate the relative transformation between pointclouds efficiently.
Due to representation in the frequency domain, PHASER enables the use of multiple input modalities.

Our evaluations show that PHASER is robust to noisy measurements, different misalignment levels, and partial overlap. 
Thus, PHASER is shown to converge to the global solution, with better performance than previous methods. These findings can have relevant implications in scenarios such as multi-robot mapping or the kidnapped robot problem.
Furthermore, we showed that our uncertainty estimate is consistent with the error of the alignment. 

As part of our future work, we would like to find synergies between the frequencies in vector and spherical space and improve our uncertainty representation by considering multiple peaks in the correlation.
Additionally, we would like to explore semantic information as a discrete random variable for the registration.
\bibliographystyle{IEEEtran}
\bibliography{bib/references.bib}

\begin{thebibliography}{10}
\providecommand{\url}[1]{#1}
\csname url@rmstyle\endcsname
\providecommand{\newblock}{\relax}
\providecommand{\bibinfo}[2]{#2}
\providecommand\BIBentrySTDinterwordspacing{\spaceskip=0pt\relax}
\providecommand\BIBentryALTinterwordstretchfactor{4}
\providecommand\BIBentryALTinterwordspacing{\spaceskip=\fontdimen2\font plus
\BIBentryALTinterwordstretchfactor\fontdimen3\font minus
  \fontdimen4\font\relax}
\providecommand\BIBforeignlanguage[2]{{%
\expandafter\ifx\csname l@#1\endcsname\relax
\typeout{** WARNING: IEEEtran.bst: No hyphenation pattern has been}%
\typeout{** loaded for the language `#1'. Using the pattern for}%
\typeout{** the default language instead.}%
\else
\language=\csname l@#1\endcsname
\fi
#2}}

\bibitem{Zhang}
J.~Zhang and S.~Singh, ``{LOAM: Lidar Odometry and Mapping in Real-time},'' in
  \emph{Proc. Robot. Sci. Syst.}, 2014.

\bibitem{Pomerleau2015}
F.~Pomerleau, F.~Colas, and R.~Siegwart, \emph{{A Review of Point Cloud
  Registration Algorithms for Mobile Robotics}}, 2015, vol.~4, no.~1.

\bibitem{Rusu2009}
R.~B. Rusu, N.~Blodow, and M.~Beetz, ``{Fast Point Feature Histograms (FPFH)
  for 3D registration},'' in \emph{Proc. IEEE Int. Conf. Robot. Autom.}\hskip
  1em plus 0.5em minus 0.4em\relax IEEE, 5 2009, pp. 3212--3217.

\bibitem{Le2019}
H.~Le, T.-T. Do, T.~Hoang, and N.-M. Cheung, ``{SDRSAC: Semidefinite-Based
  Randomized Approach for Robust Point Cloud Registration without
  Correspondences},'' \emph{Proc. IEEE Conf. Comput. Vision Pattern Recognit.},
  pp. 124--133, 4 2019.

\bibitem{Landry2019}
D.~Landry, F.~Pomerleau, and P.~Gigu{\`{e}}re, ``{CELLO-3D: Estimating the
  covariance of ICP in the real world},'' \emph{Proc. IEEE Int. Conf. Robot.
  Autom.}, vol. 2019-May, pp. 8190--8196, 2019.

\bibitem{Horn1987}
B.~K.~P. Horn, ``{Closed-form solution of absolute orientation using unit
  quaternions},'' \emph{J. Opt. Soc. Amer. A,}, vol.~4, no.~4, p. 629, 1987.

\bibitem{Magnusson2007}
M.~Magnusson, A.~Lilienthal, and T.~Duckett, ``{Scan registration for
  autonomous mining vehicles using 3D-NDT},'' \emph{Journal of Field Robotics},
  vol.~24, no.~10, pp. 803--827, 10 2007.

\bibitem{Segal2009}
A.~V. Segal, D.~Haehnel, and S.~Thrun, ``{Generalized-ICP},'' \emph{Proc.
  Robot. Sci. Syst.}, vol.~2, p. 435, 2009.

\bibitem{Tabib2018}
W.~Tabib, C.~O~Meadhra, and N.~Michael, ``{On-Manifold GMM Registration},''
  \emph{IEEE Robot. Autom. Lett.}, vol.~3, no.~4, pp. 3805--3812, 2018.

\bibitem{Parkison2019}
S.~A. Parkison, M.~Ghaffari, L.~Gan, R.~Zhang, A.~K. Ushani, and R.~M. Eustice,
  ``{Boosting Shape Registration Algorithms via Reproducing Kernel Hilbert
  Space Regularizers},'' \emph{IEEE Robot. Autom. Lett.}, vol.~4, no.~4, pp.
  4563--4570, 2019.

\bibitem{Zaganidis2018}
A.~Zaganidis, L.~Sun, T.~Duckett, and G.~Cielniak, ``{Integrating Deep Semantic
  Segmentation into 3-D Point Cloud Registration},'' \emph{IEEE Robot. Autom.
  Lett.}, vol.~3, no.~4, pp. 2942--2949, 2018.

\bibitem{Lei2017}
H.~Lei, G.~Jiang, and L.~Quan, ``{Fast Descriptors and Correspondence
  Propagation for Robust Global Point Cloud Registration},'' \emph{IEEE Trans.
  Image Process.}, vol.~26, no.~8, pp. 1--1, 2017.

\bibitem{Jiaolong2015}
J.~Yang, H.~Li, D.~Campbell, and Y.~Jia, ``{Go-ICP: A Globally Optimal Solution
  to 3D ICP Point-Set Registration},'' \emph{IEEE Trans. Pattern Anal. Mach.
  Intell.,}, vol.~38, no.~11, pp. 2241--2254, 11 2016.

\bibitem{Yang2020}
H.~Yang, J.~Shi, and L.~Carlone, ``{TEASER: Fast and Certifiable Point Cloud
  Registration},'' no.~c, pp. 1--42, 2020.

\bibitem{Bulow2013}
H.~B{\"{u}}low and A.~Birk, ``{Spectral 6DOF registration of noisy 3D range
  data with partial overlap},'' \emph{IEEE Trans. Pattern Anal. Mach.
  Intell.,}, vol.~35, no.~4, pp. 954--969, 2013.

\bibitem{Bulow2018}
------, ``{Scale-Free Registrations in 3D: 7 Degrees of Freedom with Fourier
  Mellin SOFT Transforms},'' \emph{Int. J. Comput. Vis.,}, vol. 126, no.~7, pp.
  731--750, 7 2018.

\bibitem{SrinivasaReddy1996}
B.~Srinivasa~Reddy and B.~N. Chatterji, ``{An FFT-based technique for
  translation, rotation, and scale-invariant image registration},'' \emph{IEEE
  Transactions on Image Processing}, vol.~5, no.~8, pp. 1266--1271, 1996.

\bibitem{Wang2012}
C.~Wang, X.~Jing, and C.~Zhao, ``{Local Upsampling Fourier Transform for
  accurate 2D/3D image registration},'' \emph{Computers and Electrical
  Engineering}, vol.~38, no.~5, pp. 1346--1357, 2012.

\bibitem{Makadia2006}
A.~Makadia, A.~Patterson~IV, and K.~Daniilidis, ``{Fully automatic registration
  of 3D point clouds},'' \emph{Proc. IEEE Conf. Comput. Vision Pattern
  Recognit.}, vol.~1, pp. 1297--1304, 2006.

\bibitem{Nieto2006}
J.~Nieto, T.~Bailey, and E.~Nebot, ``{Scan-SLAM: Combining EKF-SLAM and Scan
  Correlation},'' in \emph{Field and Service Robotics}.\hskip 1em plus 0.5em
  minus 0.4em\relax Springer Berlin Heidelberg, 2006, vol.~25, pp. 167--178.

\bibitem{Prakhya2015}
S.~M. Prakhya, L.~Bingbing, Y.~Rui, and W.~Lin, ``{A closed-form estimate of 3D
  ICP covariance},'' in \emph{Proc. 14th IAPR Int. Conf. Mach. Vis. Appl.}, no.
  May.\hskip 1em plus 0.5em minus 0.4em\relax IEEE, 5 2015, pp. 526--529.

\bibitem{Healy2003}
D.~M. Healy, D.~N. Rockmore, P.~J. Kostelec, and S.~Moore, ``{FFTs for the
  2-Sphere-Improvements and Variations},'' \emph{J. Fourier Anal. Appl.},
  vol.~9, no.~4, pp. 341--385, 2003.

\bibitem{Healy1996}
D.~M. Healy, D.~N. Rockmore, and S.~B. Moore, ``{FFT for the 2-sphere and
  applications},'' \emph{ICASSP, IEEE International Conference on Acoustics,
  Speech and Signal Processing - Proceedings}, vol.~3, pp. 1323--1326, 1996.

\bibitem{Driscoll1994}
J.~R. Driscoll and D.~M. Healy, ``{Computing fourier transforms and
  convolutions on the 2-sphere},'' pp. 202--250, 1994.

\bibitem{Kostelec2008}
P.~J. Kostelec and D.~N. Rockmore, ``{FFTs on the rotation group},'' \emph{J.
  Fourier Anal. Appl.}, vol.~14, no.~2, pp. 145--179, 2008.

\bibitem{Bingham1974}
C.~Bingham, ``{An antipodally symmetric distribution on the sphere},''
  \emph{The Annals of Statistics}, pp. 1201--1225, 1974.

\bibitem{Naidu2011}
V.~P. Naidu, ``{Multi-resolution image fusion by FFT},'' \emph{ICIIP 2011 -
  Proceedings: 2011 International Conference on Image Information Processing},
  no. Iciip, 2011.

\bibitem{Falk1999}
H.~Falk, ``{Prolog to A Categorization of Multiscale-Decomposition-Based Image
  Fusion Schemes with a Performance Study for a Digital Camera Application},''
  \emph{Proceedings of the IEEE}, vol.~87, no.~8, pp. 1313--1314, 1999.

\bibitem{Besl1992}
P.~J. Besl and N.~D. McKay, ``{A Method for Registration of 3-D Shapes},'' in
  \emph{Sensor Fusion IV: Control Paradigms and Data Structures}, P.~S.
  Schenker, Ed., vol. 1611, no. April 1992, 4 1992, pp. 586--606.

\bibitem{Myronenko2010}
A.~Myronenko and X.~Song, ``{Point set registration: Coherent point drifts},''
  \emph{IEEE Trans. Pattern Anal. Mach. Intell.,}, vol.~32, no.~12, pp.
  2262--2275, 2010.

\bibitem{Mellado2014}
N.~Mellado, D.~Aiger, and N.~J. Mitra, ``{Super 4PCS Fast Global Pointcloud
  Registration via Smart Indexing},'' \emph{Computer Graphics Forum}, vol.~33,
  no.~5, pp. 205--215, 8 2014.

\bibitem{Geiger2012}
A.~Geiger, P.~Lenz, and R.~Urtasun, ``{Are we ready for autonomous driving? the
  KITTI vision benchmark suite},'' \emph{Proceedings of the IEEE Computer
  Society Conference on Computer Vision and Pattern Recognition}, pp.
  3354--3361, 2012.

\bibitem{Carrillo2012}
H.~Carrillo, I.~Reid, and J.~A. Castellanos, ``{On the comparison of
  uncertainty criteria for active SLAM},'' in \emph{Proc. IEEE Int. Conf.
  Robot. Autom.}\hskip 1em plus 0.5em minus 0.4em\relax IEEE, 5 2012, pp.
  2080--2087.

\bibitem{vanicek2015geodesy}
P.~Vanicek and E.~J. Krakiwsky, \emph{{Geodesy: The Concepts}}.\hskip 1em plus
  0.5em minus 0.4em\relax Elsevier Science, 2015.

\end{thebibliography}
\end{document}